\documentclass[preprint,12pt,authoryear]{elsarticle}
\usepackage{amssymb}
\usepackage{amsmath}
\usepackage{amsthm}
\usepackage{booktabs}
\usepackage{tabularx}
\usepackage{hyperref}
\usepackage{amsfonts}
\usepackage{threeparttable} 
\usepackage{wrapfig}
\usepackage{tensor}
\usepackage{graphicx}%
\usepackage{multirow}%
\usepackage{url}
\usepackage{hyperref}
\begin{document}
\begin{frontmatter}
\title{Multi-Layered Reasoning from a Single Viewpoint for Learning See-Through Grasping} 
\author[SUSTech]{Fang Wan} 
\author[MBZUAI]{Chaoyang Song\corref{cor1}} 
\cortext[cor1]{Corresponding Email: songcy@ieee.org}
\affiliation[SUSTech]{organization={Southern University of Science and Technology},
            city={Shenzhen},
            postcode={518055}, 
            state={Guangdong},
            country={China}}


            
\affiliation[MBZUAI]{organization={Mohamed bin Zayed University of Artificial Intelligence},
            country={United Arab Emirates}}
\begin{abstract}

    Sensory substitution enables biological systems to perceive stimuli that are typically perceived by another organ, which is inspirational for physical agents. Multimodal perception of intrinsic and extrinsic interactions is critical in building an intelligent robot that learns. This study presents a Vision-based See-Through Perception (VBSeeThruP) architecture that simultaneously perceives multiple intrinsic and extrinsic modalities from a single visual input, in a markerless manner, all packed into a soft robotic finger using the Soft Polyhedral Network design. It is generally applicable to miniature vision systems placed beneath deformable networks with a see-through design, capturing real-time images of the network's physical interactions induced by contact-based events, overlaid on the visual scene of the external environment, as demonstrated in the ablation study. We present the VBSeeThruP's capability for learning reactive grasping without using external cameras or dedicated force and torque sensors on the fingertips. Using the inpainted scene and the deformation mask, we further demonstrate the multimodal performance of the VBSeeThruP architecture to simultaneously achieve various perceptions, including but not limited to scene inpainting, object detection, depth sensing, scene segmentation, masked deformation tracking, 6D force/torque sensing, and contact event detection, all within a single sensory input from the in-finger vision markerlessly. 

\end{abstract}
\end{frontmatter}
\section{Introduction}
\label{sec:Intro}

    Sensory substitution \citep{BachyRita2003SensorySubstitution} is a technique where one sensory modality is used to provide information typically perceived by another. Biological systems utilize visual cues to infer contact-based information, providing inspiration for achieving multimodal robotic perception \citep{Caspar2015Newfrontiers}. In classical settings, such as the ``Rubber Hand Experiment'' \citep{Botvinick1998RubberHands}, research has shown that humans can translate contact-based sensations into visual or auditory signals, enabling them to \textit{see} through touch \citep{Tsakiris2005RubberHand} or \textit{hear} textures \citep{Lundborg1999HearingSubstitution}. This concept naturally extends to understanding deformable interactions in both human and biological systems, where visual observations of deformations can infer underlying contact-based interactions and material properties \citep{Hang2021ModelingLearning}. This principle forms the basis of \textit{Vision-Based Deformable Perception (VBDeformP)}, particularly in robotic manipulation, where the growing adoption of miniature vision sensors \citep{Li2018SlipDetection, Alspach2019SoftBubble} and robot learning methods \citep{Lloyd2024PoseShear, Bauza2024SimPLE} are also notable. Accurately perceiving contact-based information is crucial for physical agents, such as robots, to effectively grasp and manipulate objects \citep{Mittendorfer2011HumanoidMultimodal}. Leveraging visual cues of deformations offers a promising solution \citep{Lambeta2024DigitizingTouch}. 

    Robotics research has increasingly explored vision-based techniques to understand deformable objects \citep{Weng2024InteractivePerception} and their interactions with the object-centric environment \citep{Zhu2021VisionbasedManipulation}. Simultaneously achieving a comprehensive perception of the scene (i.e., object detection, recognition, and spatial understanding) and the touch-based interactions (i.e., surface texture, material properties, contact forces) during object manipulation with robots is an active research area \citep{Bauza2024SimPLE}. 
    
    However, achieving multimodal perception from a single, compact visual input, especially when the sensor is embedded in a soft robotic fingertip, poses a significant challenge \citep{Fazeli2019SeeFeelAct}. The deformable interface between a miniature camera and the external environment is critical to the system's accuracy and applicability. Its design directly impacts the quality of both visual and contact-based information acquired, making it an emerging research area in tactile robotics \citep{Sun2022SoftThumb}. An \textit{ideal} system would be able to decouple the contact-based interaction inferred from the soft interface's deformation from the miniature vision that also perceives the scene, using a single visual input.

    \begin{figure}[h]
        \centering
        \includegraphics[width=1\linewidth]{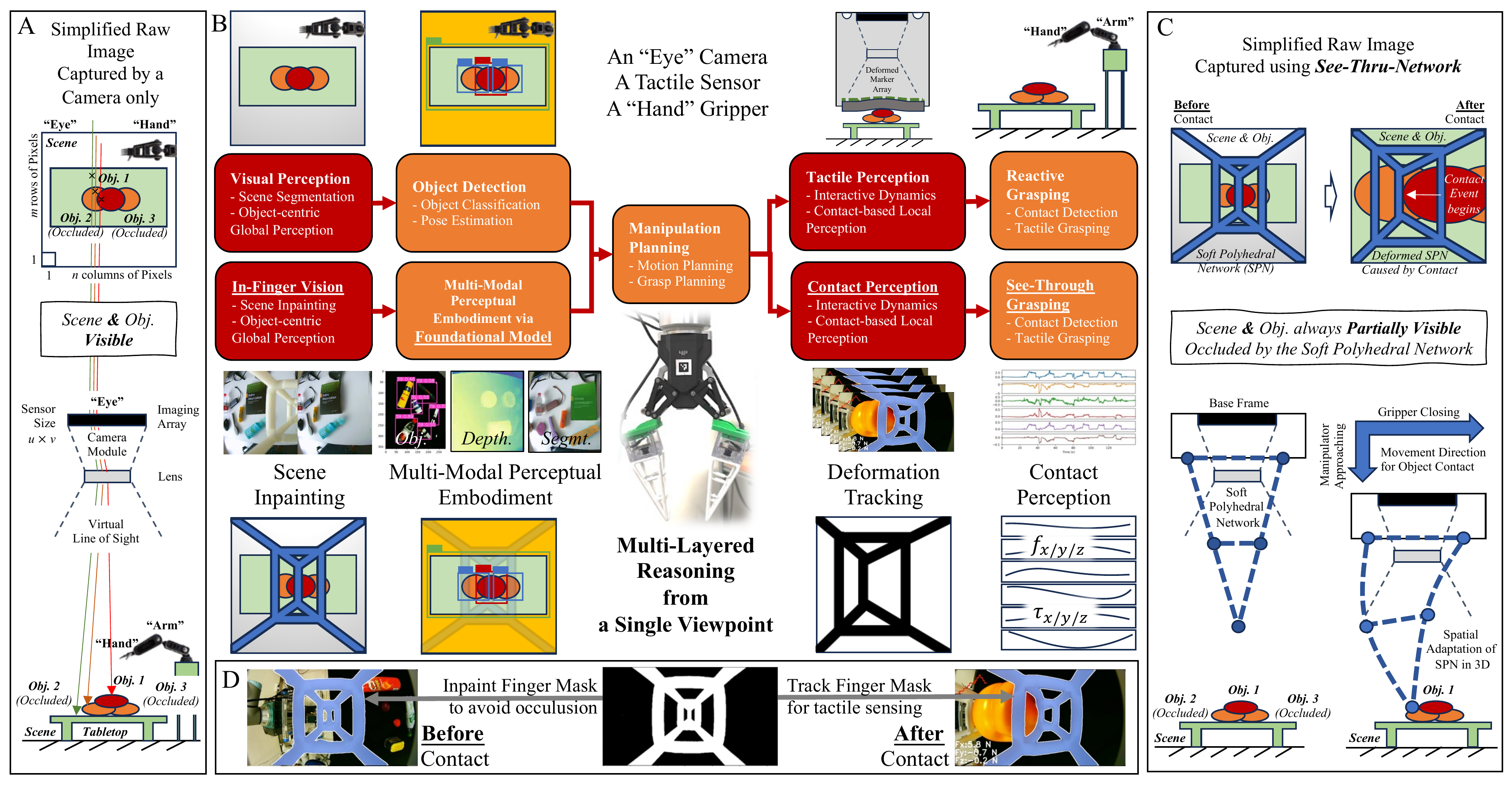}
        \caption{\textbf{Learning see-through grasping via multi-layered reasoning from a single viewpoint.}
        (A) A common setup for vision-based grasping.
        (B) Proposed pipeline in this work. 
        (C) Grasping principle via See-Thru-Network. 
        (D) Markerless representation via deformable mask tracking. 
        }
        \label{fig:Intro_PaperOverview}
    \end{figure}
    
    This work presents a \textit{Vision-based See-Through Perception (VBSeeThruP)} architecture that enables multi-layered reasoning to learn reactive grasping from a single visual input behind a deformable network, serving as a markerless representation of fingertips, as shown in Fig.~\ref{fig:Intro_PaperOverview}. The VBSeeThruP approach addresses the challenges of concurrently perceiving both finger deformation induced by contact-based interactions and rigid objects within the scene, all within a unified visual input for physical agents. For enhanced interactions with omni-directional adaptation of objects, we adopted the Soft Polyhedral Network (SPN) reported in recent research \citep{Guo2024ProprioceptiveState, Liu2024ProprioceptiveLearning}, and further achieved markerless representations through multimodal perception. The miniature vision captures real-time images of the SPN's physical deformation during contact-based interaction while preserving partial information of the external scene. Before contact, a static SPN mask is used to inpaint the occluded scene via a brief ``head-up'' camera movement. Subsequently, video object segmentation of the SPN's large-scale deformations provides markerless contact-based perception by estimating 6D forces and torques during contact. We demonstrate the viability of this approach through an ablation study and a reactive grasping task, further demonstrating its potential for object detection, depth sensing, and scene segmentation once occluded regions are inpainted. This offers a unified solution that could streamline traditional hand-eye systems into a single in-finger vision sensor. 

\section{Related Works}
\label{sec:Review}

    Vision underpins robotic perception by enabling object detection, localization, and scene understanding \citep{Kirillov2023SegmentAnything, Robinson2023RoboticVision}. However, properties such as stiffness, friction, and mass are difficult to infer solely from vision \citep{Li2024M3Tac, Zhao2025iFEM2}, calling for the integration of alternative modalities.

\subsection{Differentiating Vision-based Sensing and Perception}
\label{sec:Review-SensingVPerception}

    In robotics and cognitive science, sensing and perception are closely related but fundamentally distinct concepts, as summarized in Table \ref{tab:SensingVPerception}. 
    \begin{table}[htbp]
        \centering
        \caption{A brief comparative summary of Sensing and Perception.}
        \label{tab:SensingVPerception}
        \resizebox{\columnwidth}{!}{%
        \begin{tabular}{lll}
        \hline
        \textbf{Aspect}            & \textbf{Sensing}         & \textbf{Perception}                                  \\ \hline
        \textbf{Data Type}         & Raw, unprocessed signals & Processed, interpreted information                   \\ 
        \textbf{Role}              & Data collection          & Interpretation, understanding, and decision-making   \\ 
        \textbf{Implementation}    & Primarily hardware-level & Primarily computational or cognitive algorithms      \\
        \textbf{Complexity}        & Low-level, simpler       & High-level, complex, involving learning or inference \\ 
        \textbf{Outcome}           & Raw sensor measurements  & Meaningful representations or models                 \\ \hline
        \end{tabular}%
        }
    \end{table}
    \textit{Sensing} provides the foundational information that robots and intelligent agents need to interact with the physical world. It refers to directly acquiring raw data from the environment using sensors. It is the preliminary stage of gathering signals or measurements. It typically involves hardware (e.g., cameras, microphones, force sensors, temperature sensors); produces unprocessed, quantitative, often noisy data; and measures passive or active physical signals without interpreting their meaning. For example, a camera captures pixel-level images, and a microphone records sound waves, both of which are typically considered forms of sensing. \textit{Perception}, on the other hand, builds on sensing data, enabling robots to make informed decisions, adapt to new situations, and interact intelligently with the environment or humans. It involves processing, interpreting, and understanding sensory data to derive meaningful information or actionable insights about the environment or state. It often refers to the computational or cognitive processes involved in analyzing and interpreting sensor data, including filtering, classification, recognition, inference, and decision-making. Perception bridges the gap between raw sensing and actionable knowledge or behavior. For example, identifying objects within an image, recognizing human speech from audio signals, or estimating forces from tactile sensor data are generally considered forms of perception.

\subsection{A General Classification of Vision-based Perception}

    An object-centric representation \citep{Kroemer2021AReview} leads to rigid-body assumptions or deformable interactions for visual understanding and reasoning, promoting robotic vision via \emph{Vision-based Rigid Perception} (VBRigidP) and \emph{Vision-based Deformable Perception} (VBDeformP), which are the leading approaches.

\subsubsection{Vision-based Rigid Perception (VBRigidP)}

    VBRigidP methods traditionally assume rigid objects, leveraging tools such as Faster R-CNN \citep{Ren2015FastRCNN} or YOLO \citep{Redmon2016YOLO} for detection, combined with pose estimation techniques (PnP, ICP) \citep{Lepetit2009EPnP, Besl1992AMethod} for manipulation tasks \citep{Bohg2014DataDriven}. Although effective in structured industrial settings, VBRigidP and the rigid-object assumption become limiting for real-world scenarios involving deformable or partially deformable objects.

\subsubsection{Vision-based Deformable Perception (VBDeformP)}

    VBDeformP extends beyond rigid bodies by tracking non-rigid shapes \citep{ArriolaRios2020ModelingDeformable, Kroemer2011LearningDynamic}. Soft robotics, cloth manipulation, and surgical robotics applications exploit visual deformation cues to infer forces or material properties \citep{Zhu2022ChallengesOutlook, Deng2024GeneralPurpose}. By analyzing deformation in video, robots can estimate contact forces \citep{Luu2023SimulationLearning}, plan tasks for large-scale cloth handling \citep{Zhu2022ChallengesOutlook}, or reconstruct object geometry \citep{Guo2024ReconstructingSoft}, all without specialized tactile sensors.
    
    Emerging from VBDeformP is \textit{Vision-based Tactile Perception}, which interprets tactile interaction by analyzing deformation images \citep{Yuan2017GelSight, Li2024M3Tac}. Research has shown that detailed force and contact geometry can be inferred from images of a soft interface undergoing compression \citep{Wang2022TACTO, Yuan2017GelSight}, which can be further embedded into multi-fingered systems for object manipulation \citep{Liu2022GelSightFinRay, Zhao2023GelSightSvelte}. Recent research has demonstrated tactile perception inferred at the \textit{sensor} (e.g., normal and tangential forces \citep{Yuan2017GelSight}, vibration \citep{Kent2021WhiskSight}, thermal \citep{Li2024M3Tac}, pretouch proximity \citep{Hogan2022FingerSTS}), \textit{contact} (e.g., contact geometry \citep{Yuan2017GelSight}, force and torque \citep{Liu2024ProprioceptiveLearning}, contact events \citep{Zhang2020TowardsLearning}, material properties \citep{Yuan2017ConnectingLook}), \textit{object} (e.g., object localization \citep{Xu2024VisionBased}, shape \citep{Luu2023SimulationLearning, Duong2021LargeScale}, mass and dynamics \citep{Wang2022TACTO}, contents of containers \citep{Zhang2021DynamicModeling}), and \textit{action} (e.g., can be inferred from previous perceptual modalities for action selection and initialization, tactile feedback for low-level control, action termination, action outcome detection, and action outcome verification) levels \cite{Li2020AReview}, providing a cost-effective solution for tactile robotics \cite{Haddadin2019TactileRobots}. 
        
    It should be noted that the Vision-based Tactile Perception mentioned above is typically referred to as \textit{Vision-based Tactile Sensing (VBTS)} in the computer vision and robotics literature. However, such terminology has caused some confusion among researchers in the cognitive science field, as the raw output signals obtained are camera pixels, and the subsequent tactile information is computationally processed to infer contact-based meanings for object manipulation. In addition, it could be argued that the tactile data collected using the VBTS method is not directly measured from compression readings of force or pressure on the interaction surface, but rather processed algorithmically from raw image pixels, making it more of a perception method than a sensing method. Without losing generality, we describe this approach as a subcategory of the VBDeformP method in this work, following the general review elaborated in Sec. \ref{sec:Review-SensingVPerception}. 

\subsubsection{Vision-based See-Through Perception (VBSeeThruP)}

    In this work, we introduce Vision-based See-Through Perception (VBSeeThruP), a novel approach that combines the benefits of both VBRigidP and VBDeformP to achieve multimodal perception from a single visual input by reasoning over information from different image layers during in-finger imaging. Cognitive science research on sensory substitution inspired our research in VBSeeThruP, as we can use a single visual sensor to infer multiple perceptual modalities by reasoning across the various layers of information the camera captures. For example, our proposed VBSeeThruP architecture infers contact-related information by tracking masked deformation in the See-Thru-Network (which is not limited to the Soft Polyhedral Network design exemplified in this work). The VBSeeThruP can also infer visual-related information after scene inpainting and the use of foundational models for a rich set of perceptual modalities in an object-centric scene. The combined effect is not achievable with VBRigidP or VBDeformP alone, and it is enabled by introducing a See-Thru-Network in front of the in-finger camera. For example, when the network is 100\% filled and does not have see-through capabilities, it becomes equivalent to a VBTactileP (or VBTS) system. When it is 0\% (or fully transparent), it becomes equivalent to a VBRigidP system. 

\subsection{Multimodal Perception in Robotics with Vision}

    Combining vision and touch provides richer representations, improving classification \citep{Falco2017CrossModal}, pose estimation \citep{Murali2024SharedVisuoTactile}, and in-hand manipulation \citep{Suresh2024NeuralFeels}. This fusion is instrumental in cluttered or occluded environments \citep{Dutta2025PredictiveVisuoTactile, NavarroGuerrero2023VisuoHaptic}. Pairing visual and auditory cues boosts performance in human action classification and material recognition \citep{Pieropan2014AudioVisual, Dimiccoli2022RecognizingObject}. Multi-sensory attention mechanisms (vision, audio, touch) enhance performance for complex activities, such as pouring or packing \citep{Li2022SeeHearFeel, Wang2023RobotGaining}. Recent multimodal large language models (LLMs) combine visual perception with language-based reasoning, enabling zero-shot and few-shot policy learning \citep{Li2024ManipLLM, Jiang2023VIMA}. These systems exhibit stronger physical reasoning when additional modalities are incorporated \citep{Lai2024VisionLanguage, Driess2023PaLME}, widening opportunities for flexible, data-driven manipulation \citep{Liu2024RoboMamba}.

\section{Proposed Methods}
\label{sec:PreDis}

    Our approach begins by denoting an image \(\mathbf{X}_t \in \mathbb{R}^{m \times n}\) captured by a camera at timestep \(t\). The \(m\) and \(n\) are the user-defined output settings for the image size, measured in the total count of pixel rows and columns. Depending on the camera selection, the \(i\)th row and \(j\)th column pixel entry of the image \(x_{i,j,c,t}\) may contain the brightness intensity of a single channel for grayscale images (\(c=1\)) or three channels in red, green, and blue (RGB) for color images (\(c=1,2,3\)). Each pixel entry typically stores information in an 8-bit format, offering 256 levels from 0 to 255. Furthermore, we can add another data channel (\(c=4\)) to the RGB image to encode each pixel's transparency state.

\subsection{Markerless Design of the See-Thru-Network}
    
\begin{figure}[!h]
        \centering
        \includegraphics[width=1\linewidth]{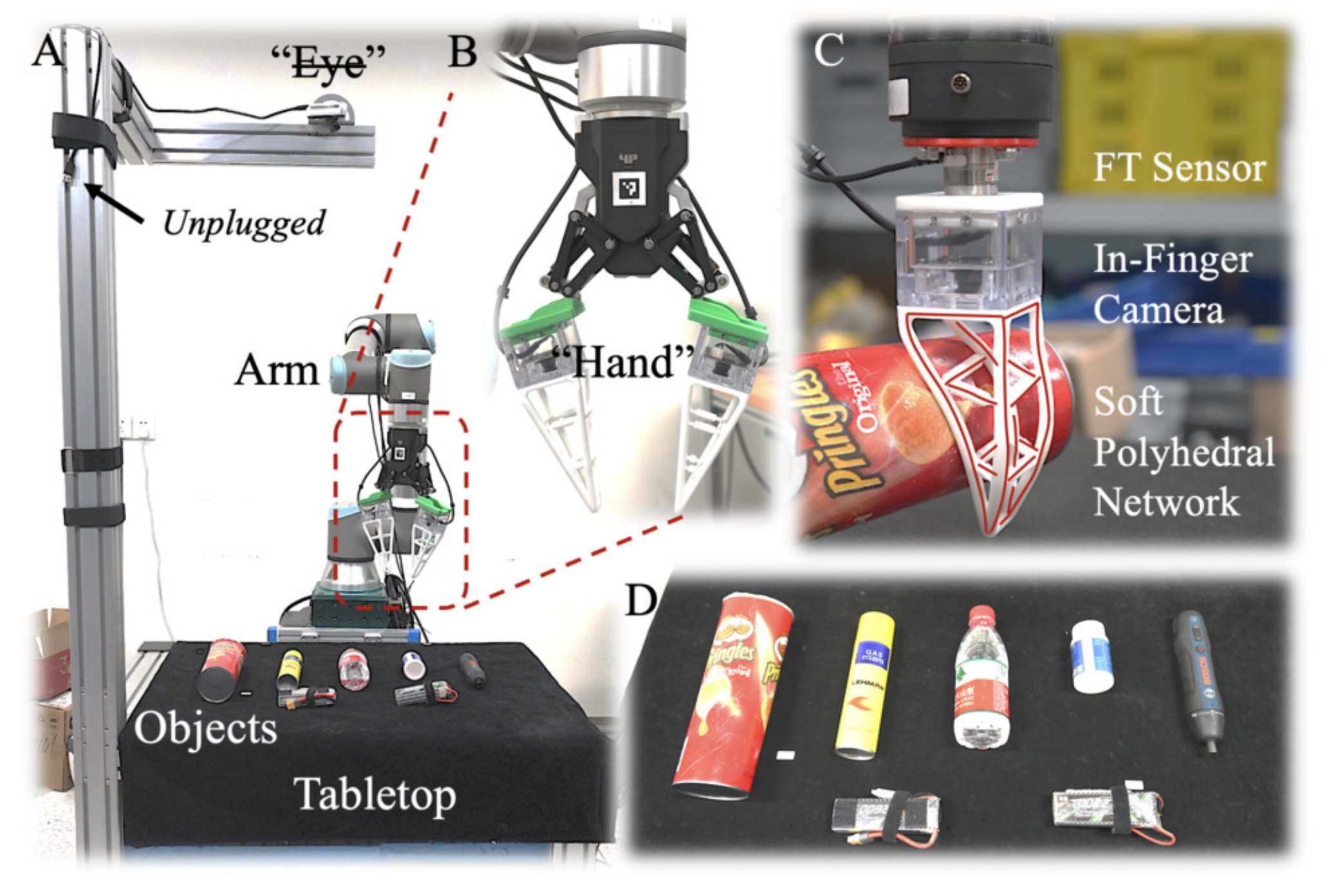}
        \caption{
            \textbf{Platform setup as a hand-eye system.}
        }
        \label{fig:Method_ExpSetup}
    \end{figure}

    Figure \ref{fig:Method_ExpSetup} is the Hand-Eye system adopted in this study, built with aluminum extrusion based on the DeepClaw workstation \citep{Wan2020Deepclaw}, housing a collaborative robot (UR10e from Universal Robots) on a pedestal and a tabletop in front covered by black fabrics. 
    Although an ``Eye'' camera is fixed on the post shown in Fig. \ref{fig:Method_ExpSetup}A, it is intentionally unplugged (therefore with a strikethrough) as we intend to implement visual perception of the scene using in-finger vision only behind a soft fingertip featuring a See-Through design using deformable materials. Figure \ref{fig:Method_ExpSetup}B shows an enlarged view of the ``Hand,'' a two-fingered gripper (Model AG-160-95 from DH-Robotics) with its rigid fingertips replaced by the soft ones. Each soft fingertip contains a soft, omni-adaptive, markerless structure based on a variant of the Soft Polyhedral Network designs \citep{Wan2022VisualLearning}, mounted on a camera housing 3D-printed by UV-curable transparent resin (Somos WaterShed XC1112), where a miniature camera (Chengyue WX605 from Weixinshijie) is fixed inside. 
    
    As shown in Fig. \ref{fig:Method_ExpSetup}C, the base plate can be customized to accommodate the installation of soft fingertips on force/torque (FT) sensors for testing or to use a gripper as a fingertip for soft and adaptive grasping. Figure \ref{fig:Method_ExpSetup}D shows the objects used for testing in this study, including a Pringles Can from the Yale-Columbia-Berkley (YCB) object sets \citep{Calli2015TheYCB} and others commonly available at research labs, including a spray can, mineral water bottles filled halfway with water, a medicine bottle, a Bosche electrical screwdriver, pouch cell. Although both soft fingertips are functional, we used only the one whose field of view partially covers the table at every frame in the following experiments.

\subsection{Formalizing Multimodal VBSeeThruP}

\begin{figure}[!h]
        \centering
        \includegraphics[width=1\linewidth]{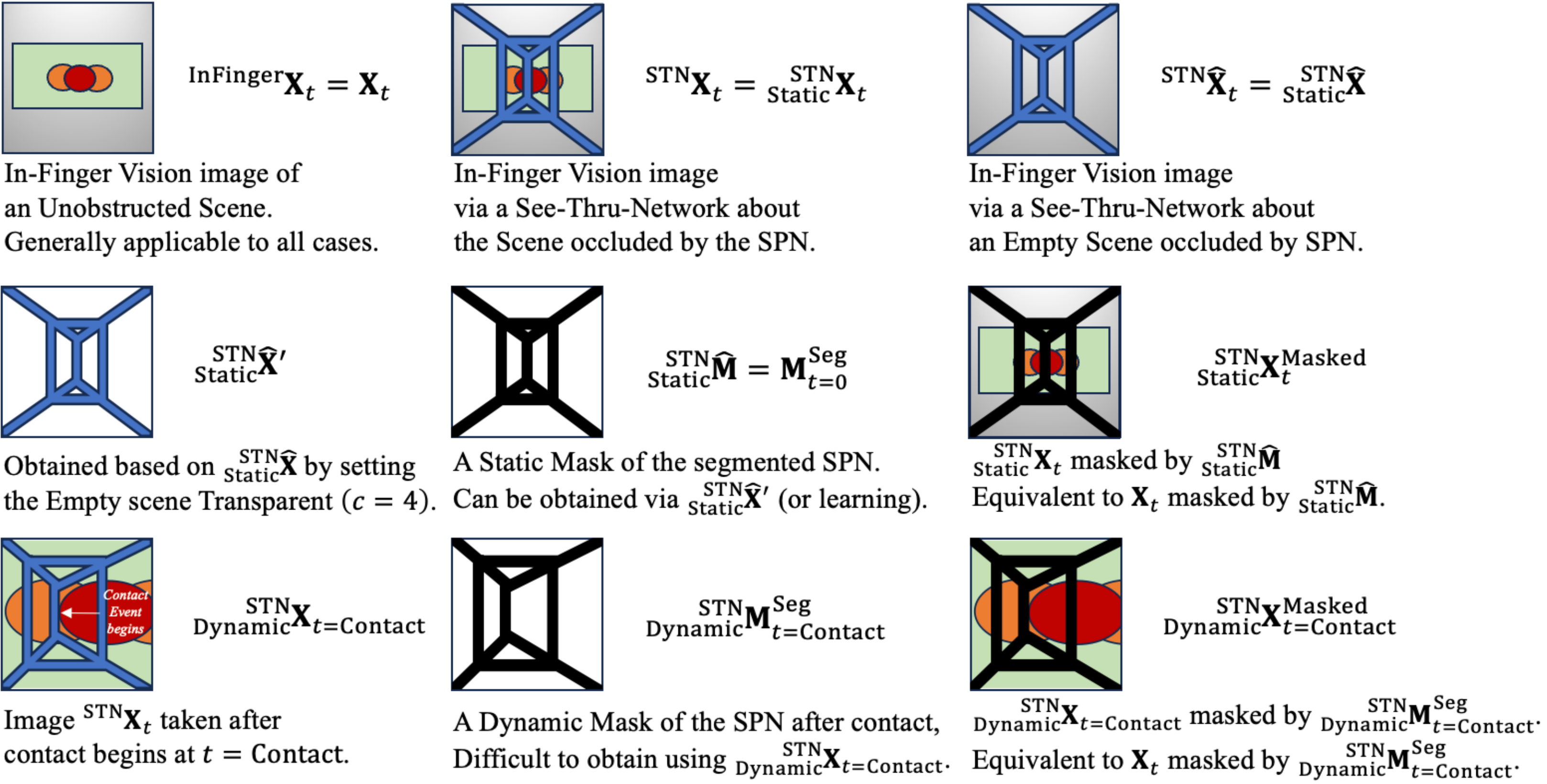}
        \caption{
            \textbf{Multi-layered reasoning for VBSeeThruP based on the SPN's See-Through design.}
        }
        \label{fig:Method_ProbForm}
    \end{figure}

    For an image \(\tensor*[^{\text{InFinger}}]{\mathbf{X}}{_t}\) taken by an in-finger vision camera, the resultant image is essentially the same as the general representation \(\tensor*[^{\text{InFinger}}]{\mathbf{X}}{_t} = \mathbf{X}_t\) reproduced in Fig. \ref{fig:Method_ProbForm}, with the ambient environment in gray. 
    Suppose the in-finger camera is used for VBDeformP with an Illuminated Chamber (IC), the pixels of the image are related to the state of the soft membrane captured within the camera's viewing range, containing zero information about the scene before contact begins \citep{Yuan2017GelSight, Li2024M3Tac}.

    However, for an image \(\tensor*[^{\text{STN}}]{\mathbf{X}}{_t}\) taken by an in-finger camera for VBSeeThruP with a See-Thru-Network (STN), the resultant image contains the scene occluded by the Soft Polyhedral Network (SPN). We denote the image as \(\tensor*[^{\text{STN}}_{\text{Static}}]{\mathbf{X}}{_t} = \tensor*[^{\text{STN}}]{\mathbf{X}}{_t}\) if the SPN is in a Static State before any contact. We can add a hat to use the symbol \(\tensor*[^{\text{STN}}_{\text{Static}}]{\mathbf{\widehat{X}}}{}\) to indicate a particular case if the scene is empty, captured as uniform pixel features. Here, the subscript timestep \(t\) is removed to simplify the representation. Since the valuable information captured in this image is purely regarding the SPN, one can add a transparency channel to obtain \(\tensor*[^{\text{STN}}_{\text{Static}}]{\mathbf{\widehat{X}}}{^{'}}\), represented in white color in Fig. \ref{fig:Method_ProbForm}. Pixel data of the SPN is preserved in this added data channel, but the pixel data of the empty scene is set to transparent. Such an operation would produce a matrix \(\tensor*[^{\text{STN}}_{\text{Static}}]{\mathbf{\widehat{M}}}{}\) that can be visualized as a Static Mask of the segmented SPN. Element-wise matrix multiplication of image \(\tensor*[^{\text{STN}}]{\mathbf{X}}{_t}\) and this Static Mask will produce masked image \(\tensor*[^{\text{STN}}_{\text{Static}}]{\mathbf{X}}{^{\text{Masked}}_{t}}\). Note that performing the same operation using raw image \(\tensor*[^{\text{InFinger}}]{\mathbf{X}}{_t}\) will produce the same masked image instead. This also explains the SPN's See-Through feature in the occlusion process for in-finger vision images. The above process generally applies when the SPN is in a Static State, meaning no deformation is induced because there is no contact event. One can use classical image processing techniques to obtain the above results.
    
    However, the above process becomes complicated when contact events occur at \(t=\text{Contact}\). On one side, the SPN enters a Dynamic State, exhibiting whole-body adaptive deformation during physical contact with the objects. On the other hand, the See-Through scene also changes during contact, further complicating segmentation. The resultant image \(\tensor*[^{\text{STN}}]{\mathbf{X}}{_t}\) captured by the in-finger camera can be represented as \(\tensor*[^{\text{STN}}_{\text{Dynamic}}]{\mathbf{X}}{_{t = \text{Contact}}}\) when contact begins. Let us assume that the Dynamic Mask of the segmented SPN, \(\tensor*[^{\text{STN}}_{\text{Dynamic}}]{\mathbf{X}}{^{\text{Seg}}_{t = \text{Contact}}}\), could be obtained. Then, one should be able to obtain a masked image of the in-finger vision, \(\tensor*[^{\text{STN}}_{\text{Dynamic}}]{\mathbf{X}}{^{\text{Masked}}_{t = \text{Contact}}}\), by performing the same pixel-wise multiplication of \(\tensor*[^{\text{STN}}_{\text{Dynamic}}]{\mathbf{X}}{^{\text{Masked}}_{t = \text{Contact}}}\) using this Dynamic Mask. Similarly, this would be equivalent to applying the Dynamic Mask over an unobstructed in-finger vision image \(\tensor*[^{\text{InFinger}}]{\mathbf{X}}{_t}\). The challenge is that a Dynamic Mask of the SPN is intricate to obtain using classical methods, especially in cluttered scenes, which often require advanced segmentation and feature engineering.

\begin{figure}[!h]
        \centering
        \includegraphics[width=1\linewidth]{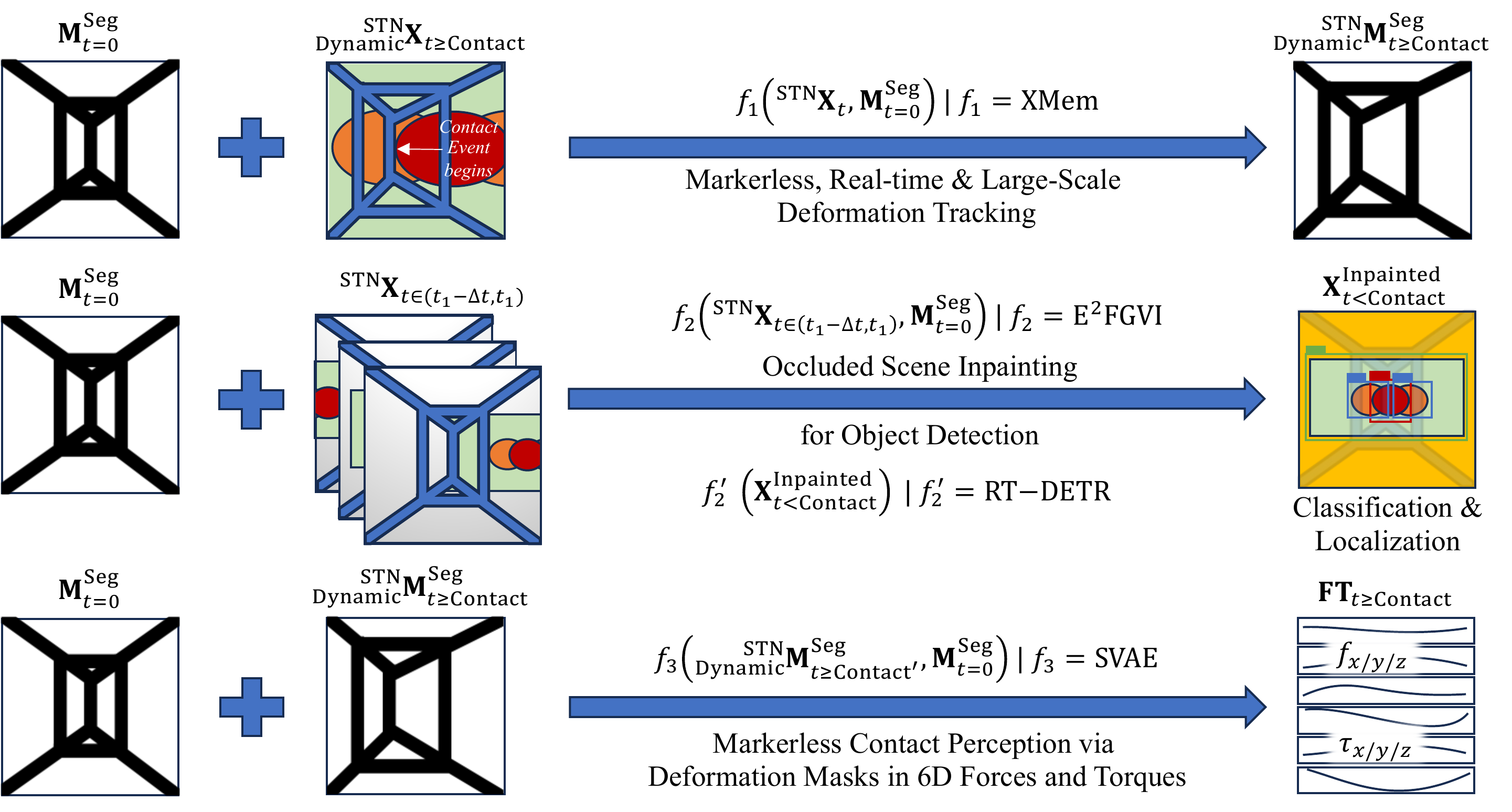}
        \caption{
            \textbf{Formalizing three research problems for VBSeeThruP via a See-Thru-Network.}
        }
        \label{fig:Method_ThreeRQs}
    \end{figure}

    Using the symbols above, we formalize three research problems of this study for a generic scenario of reactive object grasping where both 1) scene-wise visual perception for object detection and 2) object-centric force/torque sensing are considered, as shown in Fig. \ref{fig:Method_ThreeRQs} and Supplementary Video S1. 
    \begin{enumerate}
        \item Throughout the whole grasping process, can we perform markerless, real-time, and large-scale deformation tracking of the interactive medium, i.e., the Soft Polyhedral Network, in both Static and Dynamic States to obtain a pixel-wise mask \(\tensor*[]{\mathbf{M}}{^{\text{Seg}}_{t}}\)?
        \item Before contact begins, can we reconstruct the unobstructed scene \(\tensor*[]{\mathbf{X}}{_{t < \text{Contact}}}\) from occluded in-finger images \(\tensor*[^{\text{STN}}]{\mathbf{X}}{_{t < \text{Contact}}}\) for visual perception, i.e., object detection, in a Hand-Eye system via in-finger vision only?
        \item During contact-based physical interactions, can we infer high-precision 6D forces and torques \(\tensor*[]{\mathbf{FT}}{_{t \geq \text{Contact}}}\) in a markerless way?
    \end{enumerate}
    
\subsubsection{Markerless, Real-time, Large-Scale Deformation Tracking}
\label{sec:Result-Tracking}
    
    The \textit{first} problem is to enable markerless, real-time segmentation and tracking of the STN's adaptive deformation during contact, which often spans a large image area and involves large-scale spatial deformations. This problem is equivalent to video object segmentation in computer vision. A significant challenge in addressing this problem is efficiently managing the feature memory model, avoiding memory explosion while minimizing performance degradation for long-term predictions. The latest development in a Unified Memory architecture, the XMem model, exhibits state-of-the-art performance on long-video benchmarking datasets \citep{Cheng2022XMem}. 
    
\begin{figure}[!h]
        \centering
        \includegraphics[width=1\linewidth]{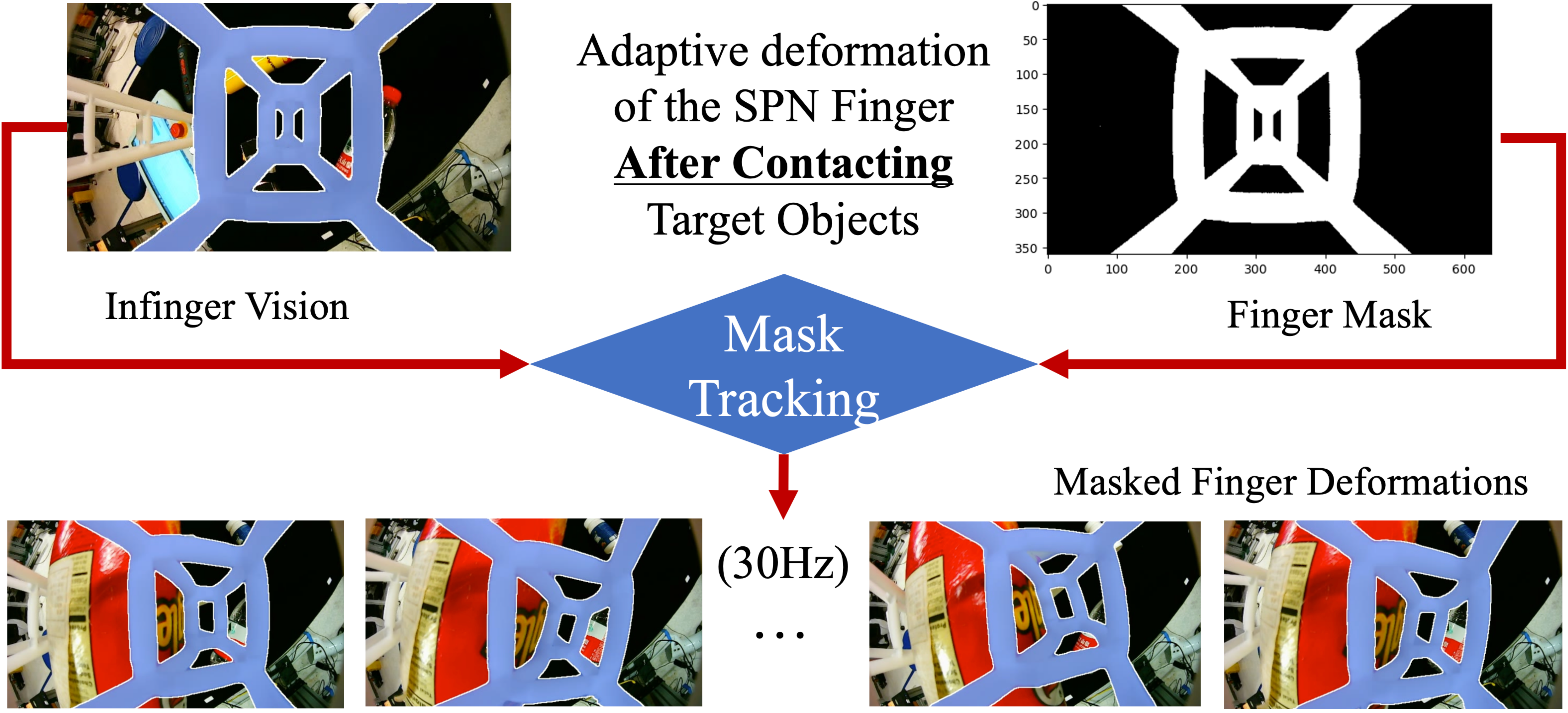}
        \caption{
            \textbf{Markerless deformation tracking of STN mask.}
        }
        \label{fig:Result_DeformationTracking}
    \end{figure}

    In this study, we experimented with the XMem model for the in-finger video stream, as shown in Fig. \ref{fig:Result_DeformationTracking}, achieving real-time (30 Hz) markerless implementation of the STN's large-scale deformation mask. 
    This is realized by using a Static Mask of the STN \(\tensor*[^{\text{STN}}_{\text{Static}}]{\mathbf{\widehat{M}}}{}\) as the initial tracking target and the in-finger vision image \(\tensor*[^{\text{STN}}_{\text{Dynamic}}]{\mathbf{X}}{_{t \geq \text{Contact}}}\) of the STN after contact. Before any contact event, this Static Mask remains the same and can be conveniently obtained manually or using advanced segmentation algorithms such as SAM \citep{Kirillov2023SegmentAnything}. The following formal representation addresses the first research problem in Fig. \ref{fig:Method_ThreeRQs}.
    \begin{equation}\label{eq:MaskTrack}
        \begin{aligned}
            \tensor*[^{\text{STN}}]{\mathbf{M}}{^{\text{Seg}}_{t}} &= f_1(\tensor*[^{\text{STN}}]{\mathbf{X}}{_t}, \tensor*[]{\mathbf{M}}{^{\text{Seg}}_{t=0}}),\\
            f_1 &= \text{XMem}.
        \end{aligned}
    \end{equation}
    We evaluated this mask segmentation on a test dataset of 1,000 samples, using the ground-truth masks as a reference. The region (J) and boundary (F) measures \citep{Perazzi2016ABenchmarkDataset} are 0.975 and 0.997, respectively, showing excellent reliability in tracking the soft finger's spatial deformation masks in a markerless way. In applications, we found that XMem's mask-tracking of the STN's deformation is not limited to Dynamic States but also applies to Static States. We successfully implemented markerless and real-time deformation tracking of the STN whenever our proposed architecture is enabled, alleviating the need for artificial markers reported in previous VBDeformP literature. This process can also be used to detect the occurrence of a contact event, as further elaborated in the Results section.

\subsubsection{In-Finger Visual Perception from Scene Inpainting}
\label{sec:Result-Inpaint}

    The \textit{second} problem is reconstructing the scene using occluded images captured by the in-finger vision for visual perception, such as object detection, providing the equivalent capability to an external ``Eye'' camera. Due to the SPN's See-Through feature, this problem becomes comparable to inpainting. Video inpainting has recently attracted significant research interest, where learning-based methods exhibit superior performances in filling in missing regions of a video with spatially and temporally coherent contents \citep{Quan2024DeepLearning}.

    \begin{figure}[!h]
        \centering
        \includegraphics[width=1\linewidth]{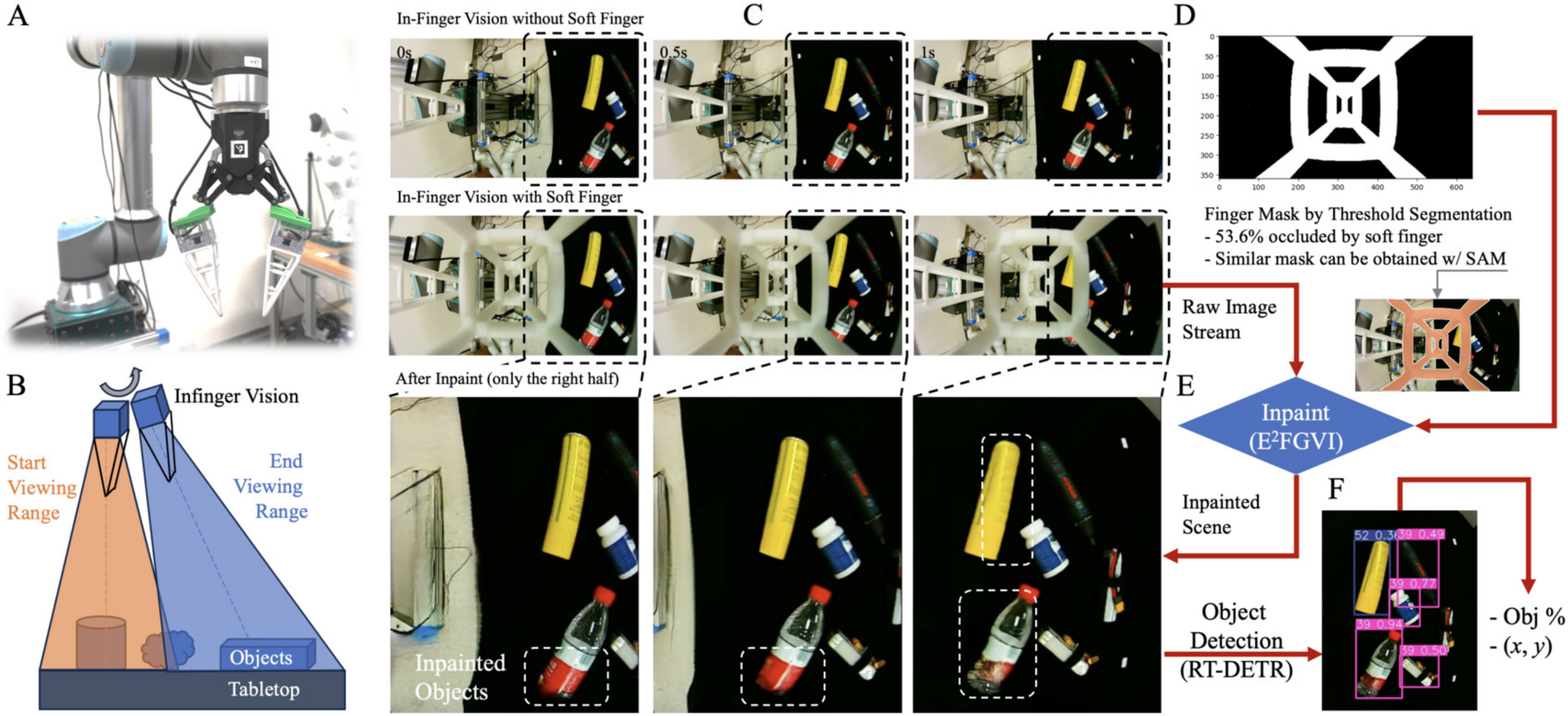}
        \caption{
            \textbf{In-finger visual perception from occluded scene inpainting using VBSeeThruP.}
        }
        \label{fig:Result_Inpainting}
    \end{figure}

    We propose moving the soft finger shown in Fig. \ref{fig:Result_Inpainting}A to obtain redundant information about the scene by scanning sequential in-finger images from different poses of the same scene, accumulating sufficient visual features from the See-Through pixels to inpaint the occluded scene. Figure \ref{fig:Result_Inpainting}B illustrates a hard-coded head-up movement of the gripper for the in-finger camera to capture a short video \(\tensor*[^{\text{STN}}]{\mathbf{X}}{_{t \in (t_1 - \Delta t, t_1)}}\) in which its viewing range sweeps across the tabletop scene. The predefined movement can be linear above the robot's working area with a fixed orientation. The first row of Fig. \ref{fig:Result_Inpainting}C shows the in-finger vision of the scene without attaching the STN, the second row shows the STN occluding the scene, and the third row shows the corresponding frames of the inpainted scene. Only the right half of the image with objects is inpainted for efficient processing. We ensured that all objects have been captured in at least one frame in the video, providing context for reconstructing frames where STN blocks objects.

    We adopted the Flow-Guided Video Inpainting (E\(^2\)FGVI) algorithm \citep{Li2022TowardsAn}, a state-of-the-art algorithm for video inpainting. To facilitate the inpainting process, we obtain a Static Mask of the STN in Fig. \ref{fig:Result_Inpainting}D using threshold segmentation against a clean background, denoted as \(\tensor*[^{\text{STN}}_{\text{Static}}]{\mathbf{\widehat{M}}}{}\). Results show that the STN occludes about 53.6\% of the in-finger image pixels. Alternatively, we can interactively obtain the same mask using the Segment Anything Model (SAM) by clicking for comparison in an actual scene \citep{Kirillov2023SegmentAnything}. Visual perception is usually performed before the contact event when \(0 \leq t < \text{Contact}\) for an object manipulation pipeline. During this period, the STN is in a Static State, meaning that the mask remains unchanged as the initial mask \(\tensor*[]{\mathbf{M}}{^{\text{Seg}}_{t=0}} = \tensor*[^{\text{STN}}_{\text{Static}}]{\mathbf{\widehat{M}}}{}\). The E\(^2\)FGVI algorithm has been reported to achieve a high Structural Similarity Index (SSI) of 0.97 on two public datasets, YouTube-VOS and DAVIS, under stationary masks. Figure~\ref{fig:Result_Inpainting}E shows the scene inpainting pipeline to implement the E\(^2\)FGVI algorithm using in-finger video clip \(\tensor*[^{\text{STN}}]{\mathbf{X}}{_{t \in (t_1 - \Delta t, t_1)}}\) and the STN's Static Mask \(\tensor*[]{\mathbf{M}}{^{\text{Seg}}_{t=0}}\) to generate inpainted scene \(\tensor*[]{\mathbf{X}}{^{\text{Inpainted}}_{t < \text{Contact}}}\) with objects on the tabletop. This inpainting process is represented as
    \begin{equation}\label{eq:Inpaint}
        \begin{aligned}
            \tensor*[]{\mathbf{X}}{^{\text{Inpainted}}_{t < \text{Contact}}} &= f_2(\tensor*[^{\text{STN}}]{\mathbf{X}}{_{t \in (t_1 - \Delta t, t_1)}},\tensor*[]{\mathbf{M}}{^{\text{Seg}}_{t=0}}),\\
            f_2 &= \text{E}^2\text{FGVI}
        \end{aligned}
    \end{equation}

    The following focuses on implementing the reactive grasping task in Fig. \ref{fig:Intro_PaperOverview}. We use the Real-Time DEtection TRansformer (RT-DETR), a state-of-the-art object detection algorithm \citep{Zhao2024DETRs}. This object detection process is represented as 
    \begin{equation}\label{eq:ObjDet}
        \begin{aligned}
            \{\%, (x,y)\} &= f'_2(\tensor*[]{\mathbf{X}}{^{\text{Inpainted}}_{t < \text{Contact}}}),\\
            f'_2 &= \text{RT-DETR}.
        \end{aligned}
    \end{equation}
    The example shown in Fig. \ref{fig:Result_Inpainting}F verifies the feasibility of visual perception using an inpainted scene from in-finger vision occluded by the SPN, producing object classification \((\%)\) and bounding box center \((x,y)\) that can be later used for grasp planning. Equations \eqref{eq:Inpaint} and \eqref{eq:ObjDet} are the formal representation of this implementation, addressing the second research problem in Fig. \ref{fig:Method_ThreeRQs}.

\subsubsection{Markerless Contact Perception via a Deformation Mask}
\label{sec:Result-Contact}

    The \textit{third} problem is to infer high-performing contact-based perception markerlessly using in-finger vision. This representation problem is usually resolved by adding artificial markers to the soft medium, which VBSeeThruP shares using either an Illuminated Chamber or a See-Thru-Network. Previous research has also explored fiducial markers within the STN, achieving high-performing proprioceptive perception, particularly when considering the STN's static and dynamic viscoelastic properties \citep{Liu2024ProprioceptiveLearning}.

    \begin{figure}[!h]
        \centering
        \includegraphics[width=1\linewidth]{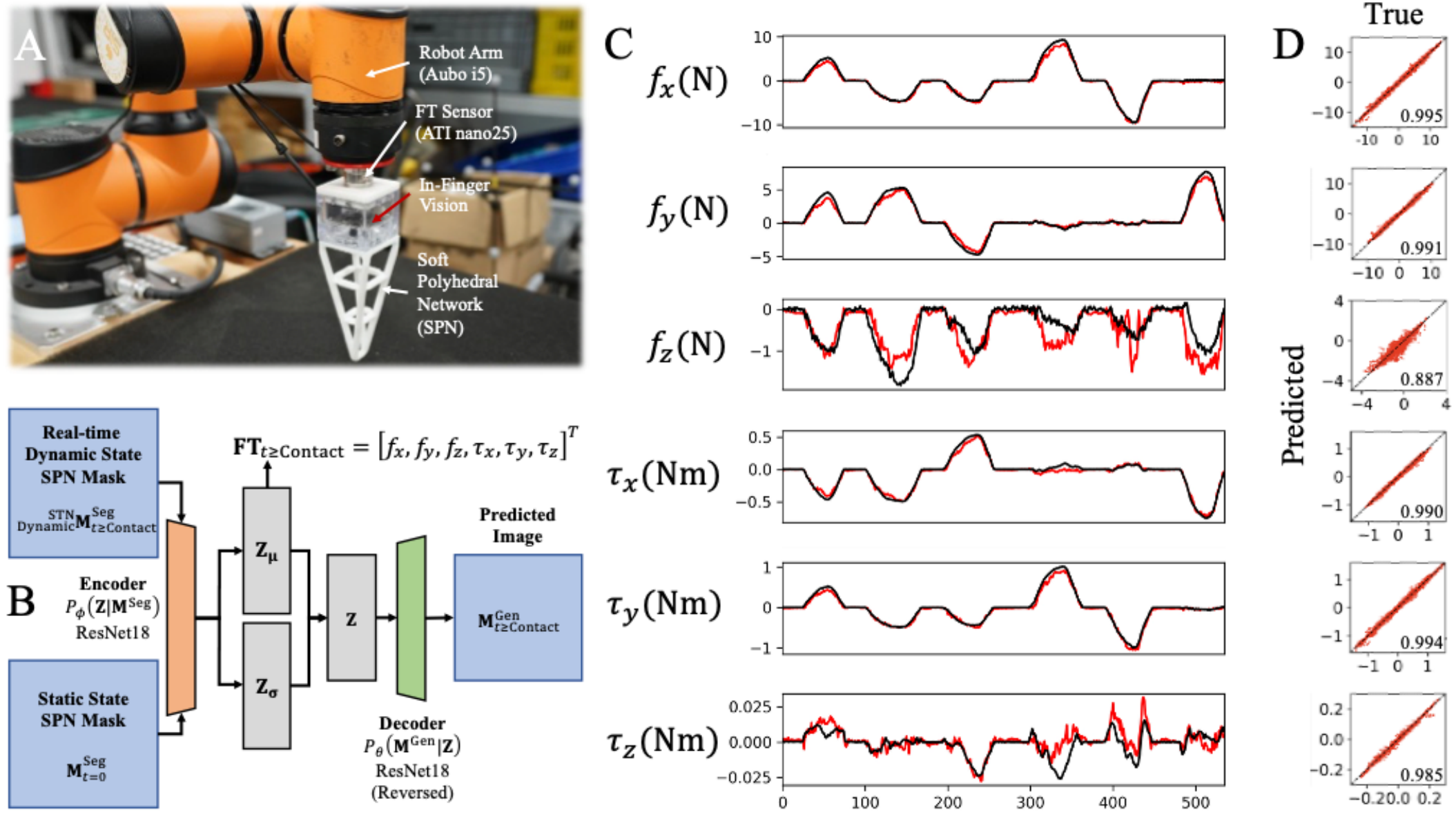}
        \caption{
            \textbf{Experiment setup, the SVAE model, and results for 6D forces and torques estimated.}
        }
        \label{fig:Result_SVAE}
    \end{figure}

    This study uses the STN masks in the Dynamic States, obtained from segmented tracking, as model inputs to infer 6D forces and torques. We adopted the Supervised Variational AutoEncoder (SVAE) for this task, and Fig. \ref{fig:Result_SVAE}A shows the experimental setup for collecting true labels of 6D forces and torques at the finger base using nano25 by ATI for calibration. The in-finger camera and FT sensor were connected to a laptop for data streaming. We collected 40,000 synchronized pairs of images and FT readings as we manually compressed the two STN fingers from various directions and contact locations. We used an 8:2 train-valid split and trained the SVAE model for 50 epochs. To avoid excessive validation scores due to potential sample dependence, we collected a separate test dataset of 1,000 samples.
    
    Figure \ref{fig:Result_SVAE}B illustrates the SVAE model. The STN masks segmented in Dynamic States \(\tensor*[^{\text{STN}}_{\text{Dynamic}}]{\mathbf{M}}{^{\text{Seg}}_{t \geq \text{Contact}}}\) and the mask template \(\tensor*[]{\mathbf{M}}{^{\text{Seg}}_{t=0}}\) are fed to the ResNet18 encoder and decoder module to reconstruct the in-finger image \(\tensor*[]{\mathbf{M}}{^{\text{Gen}}_{t \geq \text{Contact}}}\). The latent space is set to 32 dimensions. At the same time, we design an auxiliary supervised regression task of the 6D forces and torques \(\tensor*[]{\mathbf{FT}}{_{t \geq \text{Contact}}}\) simultaneously from the learned latent geometry representation. This leads to the formal representation of this implementation, addressing the third research problem in Fig. \ref{fig:Method_ThreeRQs}.
    \begin{equation}\label{eq:FTSense}
        \begin{aligned}
            \{\tensor*[]{\mathbf{FT}}{_{t \geq \text{Contact}}},\tensor*[]{\mathbf{M}}{^{\text{Gen}}_{t \geq \text{Contact}}}\} &= f_3(\tensor*[^{\text{STN}}_{\text{Dynamic}}]{\mathbf{M}}{^{\text{Seg}}_{t \geq \text{Contact}'}}, \tensor*[]{\mathbf{M}}{^{\text{Seg}}_{t=0}}),\\
            f_3 &= \text{SVAE}.
        \end{aligned}
    \end{equation}

    \begin{table}[h]
        \caption{
            \textbf{Evaluation of the Supervised Variational AutoEncoder (SVAE) model for 6D forces and torques.}
        }
        \label{tab:Result_6DFT}
        \centering
        \begin{tabular}{ccccccc}
            \hline
            \multirow{2}{*}{\textbf{MAE Datasets}} & \multicolumn{3}{c}{\textbf{Force (N)}} & \multicolumn{3}{c}{\textbf{Torque (Nm)}} \\ \cline{2-7} 
                                 & $f_x$    & $f_y$    & $f_z$   & $\tau_x$     & $\tau_y$    & $\tau_z$    \\ \hline
            Validation   & 0.241    & 0.212    & 0.14    & 0.024     & 0.027    & 0.005    \\
            Test         & 0.328    & 0.283    & 0.24    & 0.027     & 0.036    & 0.005    \\ \hline
        \end{tabular}
    \end{table}

    The overall loss comprises reconstruction loss, Mean-Squared Error (MSE) loss for force and torque prediction, and Kullback-Leibler divergence. The average time per inference is 1.9 milliseconds on an NVIDIA GeForce RTX 3080 Ti Laptop GPU. The Mean Absolute Errors of the validation and test datasets are reported in Table~\ref{tab:Result_6DFT}, indicating good generalization of the learned model. 

    Figure \ref{fig:Result_SVAE}C compares the test dataset's prediction and ground truth in time series. Figure \ref{fig:Result_SVAE}D shows the scatter plot of this comparison in the validation dataset, with excellent \(R^2\) scores above 0.985 across all components except \(f_z\). The relatively weak perforce in \(f_z\) is due to the limited adaptability along the \(z\)-axis. We also evaluate the resultant force in the \(xy\) plane and find that the mean magnitude and orientation error are 0.286 N and 2.6 degrees, respectively.

\subsection{Ablation Study}

    This section presents an ablation study demonstrating how VBSeeThruP contributes to a multimodal grasping pipeline without requiring additional or external devices. In particular, we investigate how in-finger vision and soft compliance from VBSeeThruP improve the reliability and generalization in a pick-and-place scenario. Please refer to Supplementary Video S2 for further details.

    We begin by performing depth detection of table height relative to the robot base frame. Specifically, we tilt the gripper vertically downward until it contacts the table with a force of $0.5\,\mathrm{N}$. Next, we calibrate the hand-eye in the desktop's 2D plane to obtain the perspective transformation between VBSeeThruP's in-finger camera space and the robot base frame.

    \begin{figure}[htbp]
        \centering
        \includegraphics[width=1\linewidth]{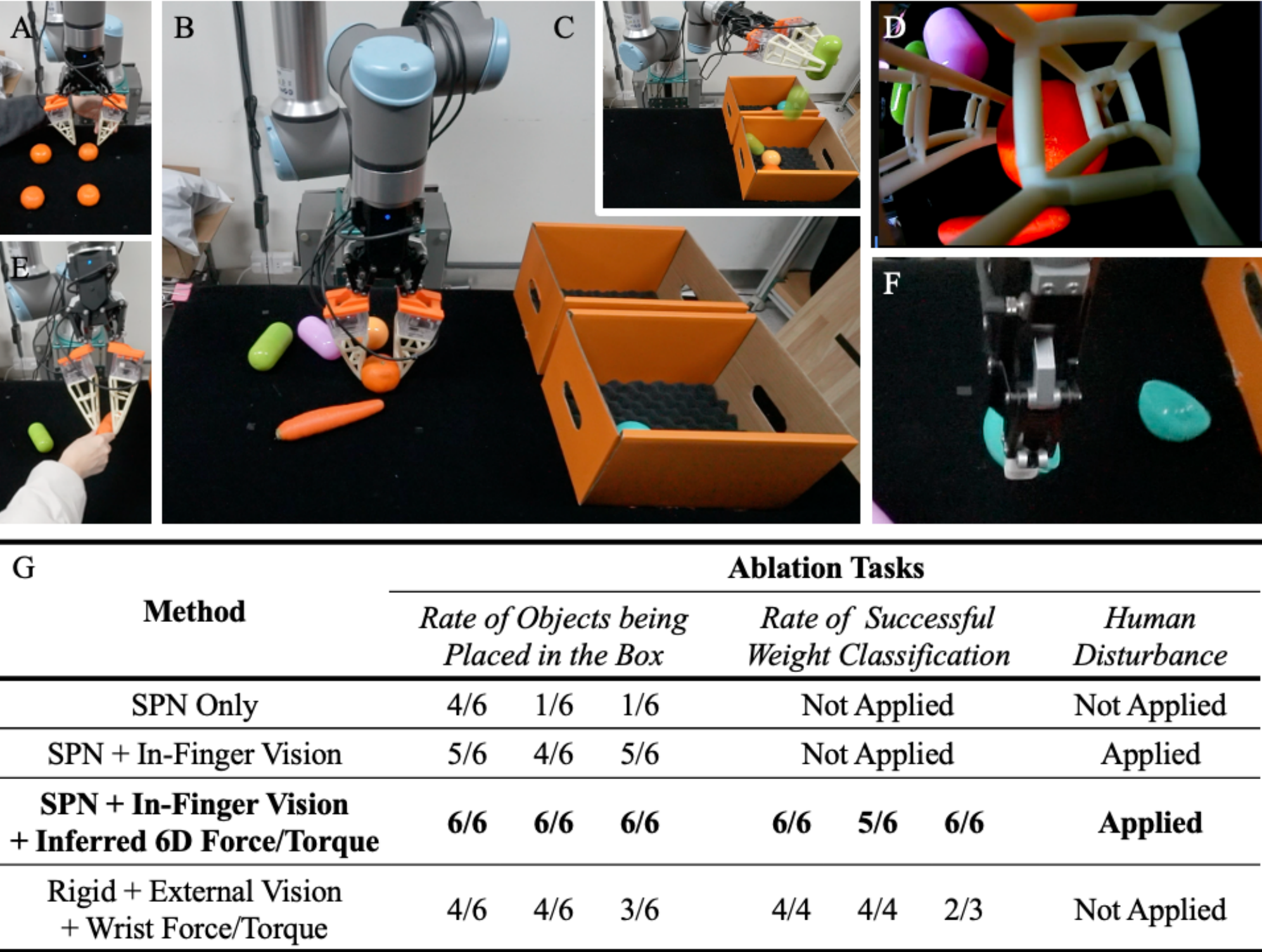}
        \caption{
            \textbf{Experiment setup and results of the ablation study.} 
            (A) Hand-eye calibration setup for perspective transformation;
            (B) Picking objects with similar shapes but different weights;
            (C) Weighing objects for sorting, then placing them into corresponding boxes;
            (D) Adaptive grasping of objects of varying sizes without specifying the fingers' closing width;
            (E) External disturbance for the carrot by intentionally removing it from the gripper;
            (F) Object damaged by a rigid finger;
            (G) Summary of the ablation study results.
        }
        \label{fig:Result_Ablation}
    \end{figure}
    
    For the calibration process, we place four oranges (or any objects of similar size) on the table and detect their centers $\{(u_i, v_i)\mid i=1,\dots,4\}$ in the image plane using YOLO11 \citep{Jocher2024UltralyticsYOLO} (see Fig. \ref{fig:Result_Ablation}A). An operator then manually bends the soft finger (for horizontal $x$-$y$ movement) to position the gripper over each orange. Once the gripper is centered above an orange, we record the corresponding world coordinates $\{(x_i, y_i)\mid i=1,\dots,4\}$ in the robot base frame by twisting the soft finger (clockwise and counter-clockwise to mark the start and end of data recording). The overall setup is shown in Figs. \ref{fig:Result_Ablation}B\&C, and a full demonstration is provided in the supplementary video. Notably, our method solely relies on VBSeeThruP for depth detection, whereas a typical solution would require an external depth camera.
    
    We employ the following control policy to facilitate intuitive robot teaching:
    \begin{equation}\label{eq:TeachRobot1}
        v_i = 
        \begin{cases}
            0, & |f_i| < 1,\\
            V_{\max}\,(f_i - 1)/4, & 1 \le |f_i| \le 5, \quad i \in \{x, y\},\\
            V_{\max}, & |f_i| > 5,
        \end{cases}
    \end{equation}
    where $v_i$ denotes the translational velocity in the $i$-th direction, and $f_i$ the corresponding finger deformation. For convenience, we also define two additional commands triggered by the torque $\tau_z$:
    \begin{equation}\label{eq:TeachRobot2}
        \text{command} = 
        \begin{cases}
            \text{Record current position}, & \tau_z > 0.2,\\
            \text{Exit program}, & \tau_z < -0.2.
        \end{cases}
    \end{equation}
    Using the pairs of recorded images and robot-world coordinates, we compute the homography matrix $H$, enabling transformations between the in-finger vision space and the robot base frame.

\subsubsection{Experiment Setup}

    We conducted a pick-and-place sorting task with six objects sharing similar geometry but different weights, including 1) Two hollow plastic balls ($6\,\mathrm{cm}$ radius): one empty and one filled with a $100\,\mathrm{g}$ metal block; 2) Two plastic capsules ($5\,\mathrm{cm}$ radius, $10\,\mathrm{cm}$ length): one empty and one containing $100\,\mathrm{g}$ of jelly balls; 3) One orange ($6.7\,\mathrm{cm}$ radius, $131\,\mathrm{g}$); and 4) One carrot ($3.1\,\mathrm{cm}$ radius, $122\,\mathrm{g}$).
    
    We tested four different ablation conditions:
    \begin{enumerate}
        \item \textbf{SPN Only}: We use the SPN's adaptive grasping with no in-finger vision or FT. The grasp position $(x,y)$ and rotation $r$ are randomly generated, and the gripper's closing width is fixed at $1.5\,\mathrm{cm}$ for all objects.
        \item \textbf{SPN + In-finger Vision}: We augment the SPN with soft fingers' in-finger vision for scene inpainting and YOLO11-based \citep{Jocher2024UltralyticsYOLO} object localization. The grasp width remains fixed at $1.5\,\mathrm{cm}$.
        \item \textbf{SPN + In-finger Vision + Inferred 6D FT}: Besides in-finger vision, the system leverages the soft finger's deformable sensor for estimating 6D forces and torques. We use reactive grasp control to maintain a constant grasp force of $3\,\mathrm{N}$ ($f_x=3\,\mathrm{N}$) and continuously monitor grasp force during transport. If the force drops to zero (e.g., due to an external disturbance), the system will attempt to regrasp.
        \item \textbf{Rigid + External Vision + Wrist FT}: We replace soft fingers with the original rigid fingertips and rely on an external camera for object localization. The gripper width is set based on the object's oriented bounding box, and wrist-mounted force/torque sensing on the UR10e provides feedback.
    \end{enumerate}
    
    We evaluate each setup on three aspects:
    \begin{itemize}
        \item \textbf{Pick-and-place success}: The ability to pick objects from the tabletop and place them into a designated box.
        \item \textbf{Weight classification before placement}: In the third ablation condition, we aggregate the two $f_y$ readings from each soft finger and estimate the object mass $m_{\mathrm{obj}}$ as
        \begin{equation}\label{eq:Weight}
            m_{\mathrm{obj}} = \frac{|f^1_y| + |f^2_y|}{9.8}.
        \end{equation}
        We use the $f_z$ reading from UR10e's wrist FT sensing in the fourth ablation condition.
        \item \textbf{Recovery from human disturbance}: We manually remove the carrot from the gripper during the transport (Fig. \ref{fig:Result_Ablation}D) to test whether the robot detects the event and regrasps successfully.
    \end{itemize}

\subsubsection{Results and Discussion}

    Each experiment scenario was repeated three times. We observed that even in the most straightforward configuration (\textbf{SPN Only}), objects could often be picked up without damaging them or triggering safety stops, thanks to the soft finger's compliance. For example, the soft, deformable fingertips are easily adapted to objects from different angles, as shown in Fig. \ref{fig:Result_Ablation}E (grasping an orange from the side with bending and twisting deformations). In contrast, using the original rigid fingertips without carefully specifying object dimensions often resulted in collisions and emergency stops.
    
    When \textbf{in-finger vision} was added (\textbf{SPN + In-finger Vision}), the grasp location $(x,y)$ and orientation $r$ were derived from YOLO11, significantly improving the success rate over random grasps. Nevertheless, the lack of force feedback made the system less robust to disturbances, causing the robot to fail regrasp attempts in several trials (as shown in the supplementary video). In one case, the fingertip was misaligned and exerted excessive force on a plastic ball, causing it to slide out of the gripper.
    
    The configuration with \textbf{SPN + In-finger Vision + Inferred 6D FT} overcame these issues using 6D force/torque estimates for closed-loop grasping. As a result, it achieved a 100\% success rate under the same disturbance conditions. It also enabled weight classification to differentiate between empty balls and allowed the system to adaptively regrasp objects when an object that had fallen was detected.
    
    Comparatively, the \textbf{Rigid + External Vision + Wrist FT} setup offered an unobstructed camera view of the workspace but lacked the compliance of soft fingers. Small errors in object detection or hand-eye calibration frequently led to inappropriate grasp forces, sometimes resulting in damaged objects (Fig. \ref{fig:Result_Ablation}F) or failed grasps. All experiment outcomes are summarized in Fig. \ref{fig:Result_Ablation}G.

    The results indicate that VBSeeThruP's in-finger vision and compliance significantly improve the success rate and robustness of pick-and-place tasks over alternative setups. The robot achieves reliable picking, weight classification, and effective regrasping after disturbances without requiring external cameras or specialized sensors. Even when paired with an unobstructed external vision system, rigid fingers are more likely to cause failure or exert excessive force, underscoring the advantage of soft fingers with built-in sensing. 

\section{Results and Discussion}
\label{sec:Exp}

\subsection{Learning Reactive See-Through Grasping}

    This section demonstrates the application of the proposed architecture for learning reactive see-through grasping, shown in Fig. \ref{fig:Intro_PaperOverview}, in an object-picking task, shown in Fig. \ref{fig:Method_ExpSetup}, using VBSeeThruP with a See-Thru-Network to achieve object detection and 6D forces and torques for reactive grasping simultaneously. Please refer to Supplementary Video S3 for a video demonstration.
    
    The task begins with visual perception of the scene using the VBSeeThruP for object classification and localization. During the first attempt, the gripper will pick up the object at the center of the bounding box, with a fixed rotation angle \(\theta_0\) and a depth \(z_0\) perpendicular to the table. We intentionally used a simpler algorithm without the optimal orientation to test reactive grasping. 
    \begin{figure}[!h]
        \centering
        \includegraphics[width=1\linewidth]{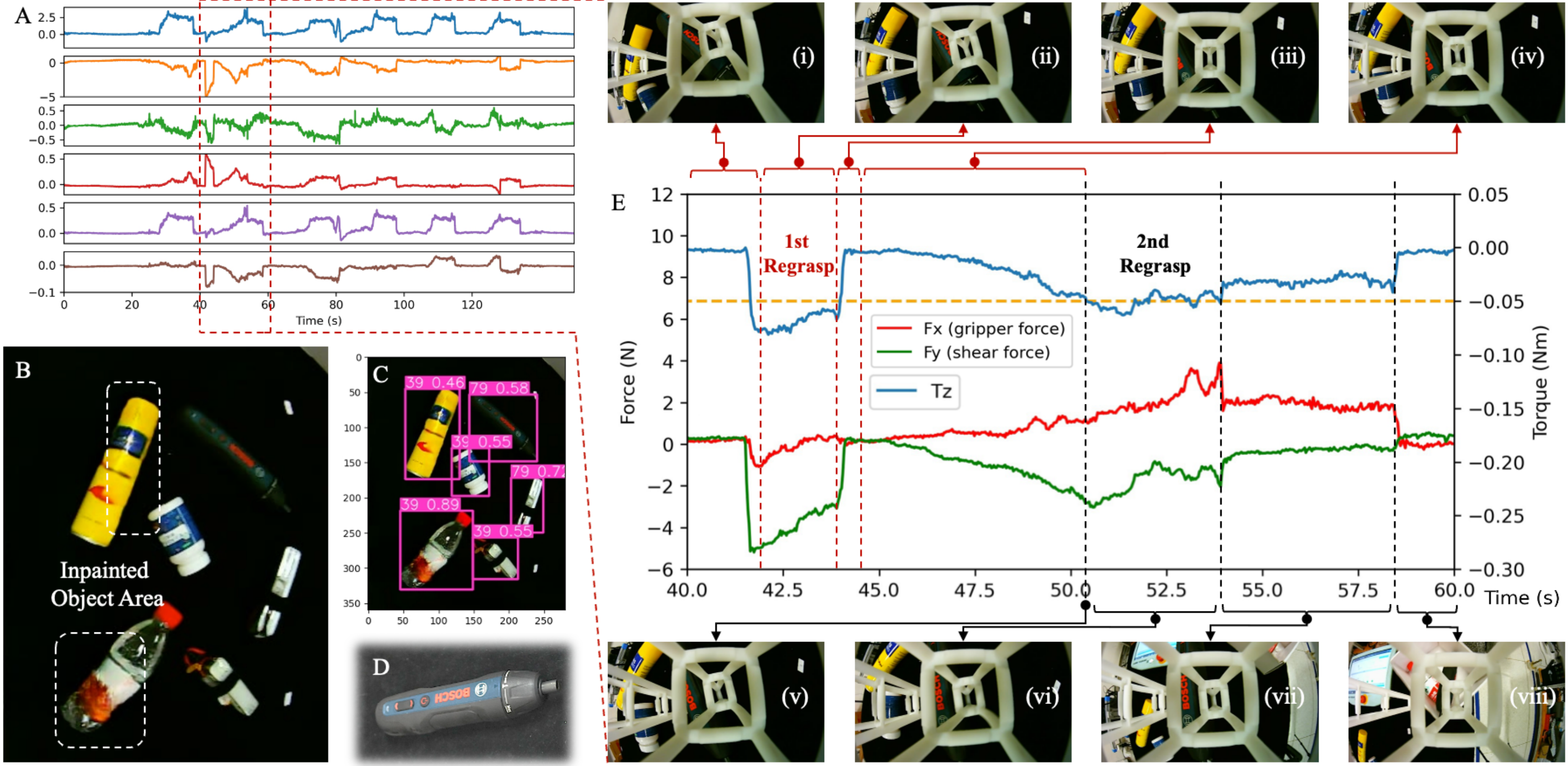}
        \caption{
            \textbf{VBSeeThruP demonstration for learning reactive see-through grasping.} 
            (A) Time series 6D forces and torques were recorded, where the red dashed box marks the reactive grasping process for the Bosche electrical screwdriver.
            (B) Inpainted scene via in-finger vision, where the white dashed boxes are the inpainted portions of the scene and objects. 
            (C) Visual perception of the inpainted scene via object detection. 
            (D) The Bosche electrical screwdriver that requires reactive grasping. 
            (E) A detailed plot of the 6D forces and torques for reactive grasping of the Bosche electrical screwdriver and multiple screenshots of in-finger vision.
        }
        \label{fig:Result_ReactiveGrasping}
    \end{figure}

    While approaching the initial grasping pose, VBSeeThruP detects if it has contact with the object by inferring real-time 6D forces and torques based on the finger's whole-body deformation. Due to the network design, the finger can deform in a twisted manner, generating torque about the \(z\)-axis. When \(\tau_z\) is detected to be larger than 0.05 Nm, the object orientation is severely different from the gripper's initial pose, suggesting a less ideal grasping that can be optimized through regrasping by turning the wrist joint reversely to reduce \(\tau_z\) for an enhanced adaptation and grasping robustness. When \(\tau_z\) is within the threshold, the gripper closes its fingers until the gripping force \(f_x\) reaches the predefined maximum gripping force of 3 N. Then, the gripper lifts the object and drops it into the placement box. We repeated the process until all five objects were cleared from the table. 

    Figure \ref{fig:Result_ReactiveGrasping}A shows the recorded 6D FT sensing data using VBSeeThruP for the table-cleaning task, where the six peaks in \(f_x\) indicate the grasping of the six test objects. The inpainted scene using VBSeeThruP is shown in Fig. \ref{fig:Result_ReactiveGrasping}B, where the occluded regions are inpainted with coherent details, and the overall contours of the objects are well reconstructed. Although some surface textures generated are blurry, they are sufficiently accurate to reflect their actual textures. Figure \ref{fig:Result_ReactiveGrasping}C shows object detection results based on the inpainted scene, showing bounding boxes for each detected object. The most exciting result is the regrasping of the Bosche electrical screwdriver shown in Fig. \ref{fig:Result_ReactiveGrasping}D. By inspecting the detailed FT history in Fig. \ref{fig:Result_ReactiveGrasping}A, we identified two regrasping attempts of the Bosche electrical screwdriver performed by the VBSeeThruP with the soft fingers. 

    Figure \ref{fig:Result_ReactiveGrasping}E is an enlarged view of the red dashed box region in Fig. \ref{fig:Result_ReactiveGrasping}A. When we performed the first grasping attempt in Fig. \ref{fig:Result_ReactiveGrasping}E(i), the finger detected the \(\tau_z\) greater than 0.05 Nm in the negative direction. Then, in Fig. \ref{fig:Result_ReactiveGrasping}E(ii), the first reactive grasping begins rotating the gripper about the \(z\)-axis to reduce \(\tau_z\). During the process, the screwdriver suddenly turns to another side and temporarily loses contact with the finger in Fig. \ref{fig:Result_ReactiveGrasping}E(iii), as both \(f_x\) and \(f_y\) dropped to 0 suddenly. In Fig. \ref{fig:Result_ReactiveGrasping}E(iv), the gripper keeps closing, attempting to detect contact with the screwdriver again to keep increasing \(\tau_z\) for an enhanced grasping pose. Next, the soft finger detected contact with the screwdriver again in Fig. \ref{fig:Result_ReactiveGrasping}E(v), marking the starting point of the second reactive grasping attempt. The soft finger repeated the same procedure by performing the second reactive grasping in Fig. \ref{fig:Result_ReactiveGrasping}E(vi), rotating the gripper about the \(z\)-axis until the detected \(\tau_z\) is reduced towards 0. As a result, the screwdriver is secured between the fingers and moved toward the drop box in Fig. \ref{fig:Result_ReactiveGrasping}E(vii). Finally, the soft finger dropped the screwdriver with forces and torques approaching zero, as shown in Fig. \ref{fig:Result_ReactiveGrasping}E(viii).

\subsection{Towards a Multimodal Soft Touch vis VBSeeThruP}

    Here, we discuss the extended modalities perceivable via our proposed architecture in Fig. \ref{fig:Intro_PaperOverview}, providing an intuitive understanding of the three research problems illustrated in Fig. \ref{fig:Method_ThreeRQs} that aim to achieve multimodal perception with a soft touch. Please refer to the following sections and Supplementary Video S4 for further details.

    \begin{figure}[!h]
        \centering
        \includegraphics[width=1\linewidth]{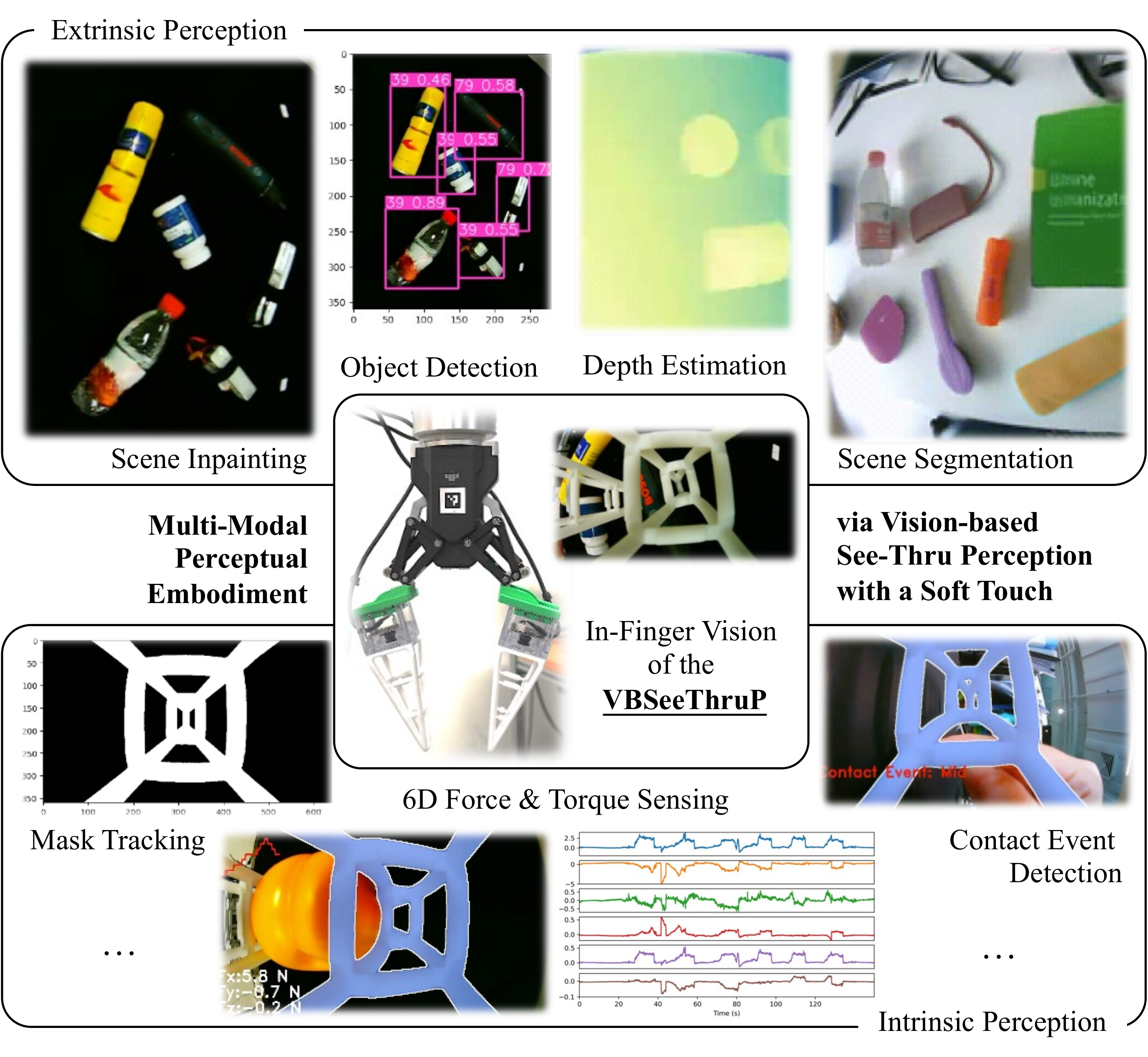}
        \caption{
            \textbf{Towards multimodal perceptual embodiment.}
        }
        \label{fig:Discuss_MultiModal}
    \end{figure}

    Using a single sensory input from the in-finger vision, featuring See-Thru-Networks, we present a multimodal approach to robotic perceptual embodiment via VBSeeThruP with soft touch, as shown in Fig. \ref{fig:Discuss_MultiModal}. Our proposed architecture can be generalized to further generate various outputs for extrinsic perception, including inpainted scenes, object detection, depth sensing, and scene segmentation. For intrinsic perceptions, we demonstrate the implementation of our architecture, which generates outputs including masked deformation tracking, real-time, high-accuracy 6D FT sensing, and contact event detection. Further expansion to other intrinsic or extrinsic modalities can also be implemented. 

\subsubsection{Markerless, Real-Time \& Large-Scale Deformation Tracking}

    VBSeeThruP generically describes a class of soft robotic fingers/fingertips that involves a See-Thru-Network with in-finger vision as a design feature. The See-Through feature enables the in-finger vision to simultaneously capture visual features from the finger structure and the external environment. In this study, the See-Thru-Network is implemented using the Soft Polyhedral Network. However, researchers have a wide range of design choices to customize the soft interface as needed, provided the resultant image contains sufficient details from both the finger structure and the external environment. The resultant performance is determined by the level of detail when extracting the soft structure's deformation and the reusable information left to reconstruct the scene. 

    Algorithmic development in video object segmentation is a well-researched field that enables convenient implementation of this capability for See-Thru-Networks with in-finger vision. The 30 Hz mask tracking implemented in this study is sufficiently high to enable real-time processing for robotics without adding to the computational burden of the robotic systems. Future research to enable higher mask-tracking bandwidth is preferred to increase the upper bound for collecting sensory feedback (towards 1 kHz) from in-finger vision, requiring higher-framerate miniature cameras and high-performance mask-tracking algorithms for more challenging robotic applications. The quality of mask segmentation at this step generally determines the maximum performance one can expect from the VBSeeThruP architecture in multimodal perception. The solution to this problem provides the foundation for implementing suitable robotic solutions to address the second and third research problems.

\subsubsection{Multimodal Visual Perception of the Occluded Scene via In-Finger Vision}

    An intuitive understanding of the second research problem is whether we can utilize in-finger vision for environmental perception, which is inherently multimodal, thereby eliminating the need for an external camera in modern robot workstations.
    
    Using in-finger vision for environmental perception is counterintuitive. Having an \textit{Eye} camera mounted on the robot base is generally perceived as bio-inspired by humans or other living creatures, where visual organs are located on the body for environmental perception. This is also a feature shared by most robotic systems. After masked segmentation of the See-Thru-Network, the original image becomes fragmented. Still, it could be alternatively \textit{repaired} using inpainting, a highly active field of research in computer vision and widely adopted practice throughout history. A direct result of inpainting the in-finger vision is the reconstruction of the scene, producing visual perceptual capabilities \textit{almost} as good as those of a typical Eye camera. It should be noted that the inpainting technique is computationally intensive, requiring high-performing hardware and long processing time, neither of which is ideal for robotic applications. In this study, we added an extra head-up motion during motion planning to collect an extra 1-second video clip for inpainting as a trade-off.

    In the sections above, we demonstrated object detection using in-finger vision inpainted images. Here, we showcase the performance of achieving other modalities, such as depth sensing and scene segmentation, using the inpainted images, which are also widely used in modern robotics. In principle, robotists can reuse the inpainted images for any desired extra modalities, just as in a computer vision problem, making the VBSeeThruP architecture a generic solution for robotics, where constraints on hardware cost and computational power are significant considerations.

    \begin{figure}[!h]
        \centering
        \includegraphics[width=0.75\linewidth]{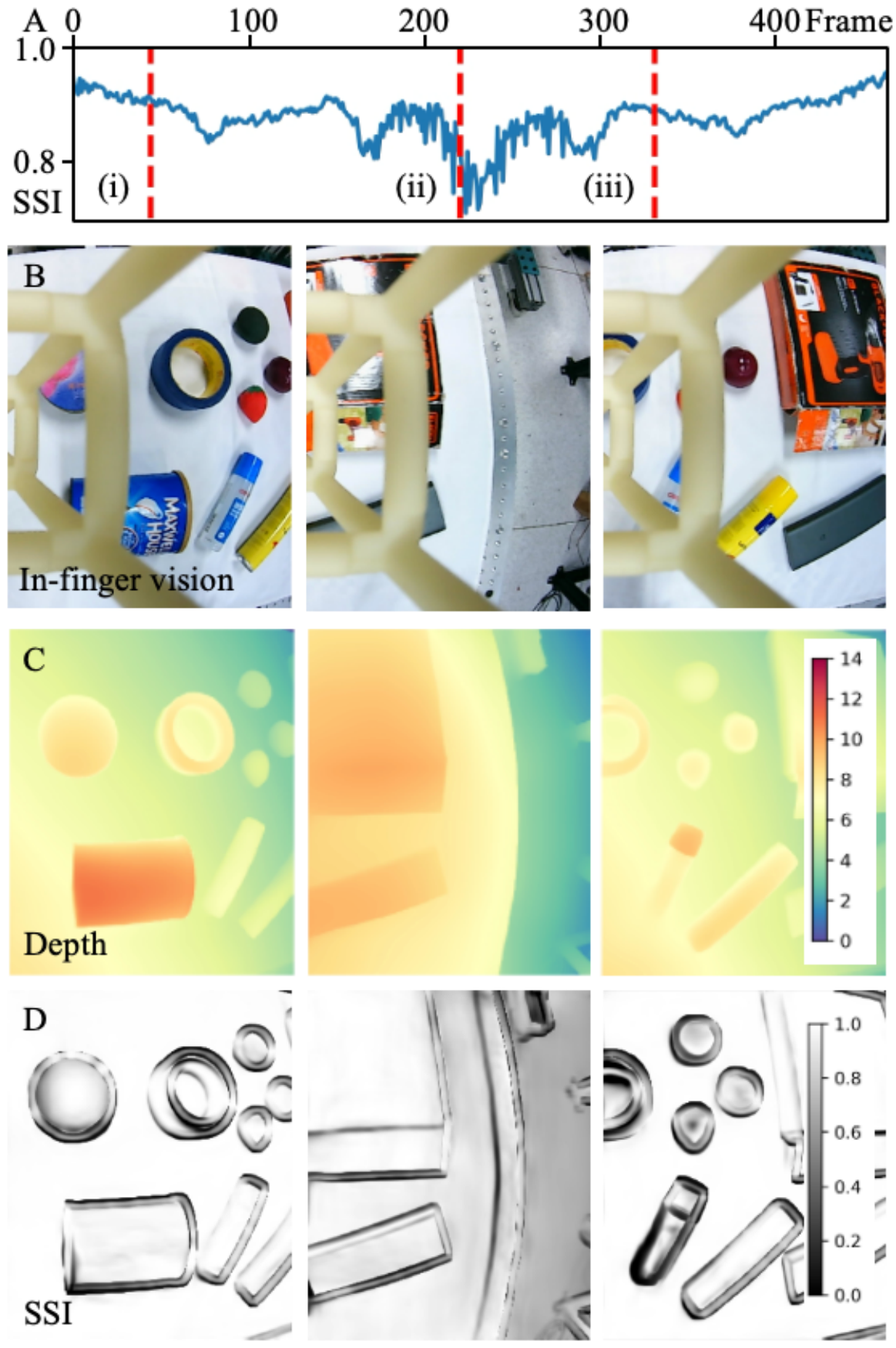}
        \caption{
            \textbf{Depth estimation by seeing through the finger after inpainting.} 
            (A) Time series of Structural similarity index (SSI) measures between depth estimations from the inpainted and unobstructed views. The red lines marked in (i), (ii), and (iii) indicate the snapshot frames of (B) in-finger vision, (C) reconstructed depth map, and (D) SSI map.
        }
        \label{fig:Discuss_Depth}
    \end{figure}

    \paragraph{Depth Sensing} We estimated monocular depth from the inpainted scene as a proof-of-concept. We compared the reconstructed scene depths from both the in-finger camera and a camera with an unobstructed view of the robot's working space. In this experiment, ten objects with various shapes and colors (milk powder can, tape, rubber ball, strawberry, glue, plum, glue stick, power drill with a packaging box, and drill toolkit) were placed on the tabletop. The robotic arm swept across the tabletop from left to right and returned to the starting point with a fixed orientation while the in-finger camera inspected the objects. We recorded both the in-finger and unobstructed videos by removing the SPN from the gripper. The former was further inpainted by the E\(^2\)FGVI algorithm. The inpainted and unobstructed videos are then fed to the Depth Anything V2 model \citep{Yang2024DepthAnythingV2} to obtain the depth estimations. Pixel-wise estimations provide relative depth values, offering helpful information about the relative poses of objects on the table. Hence, instead of comparing the actual depth map values, we use the structural similarity index (SSI) to evaluate the similarity between the depth map reconstructed from the inpainted view and that from the unobstructed view. 
    
    As shown in Fig. \ref{fig:Discuss_Depth}A, the index varies between 0.71 and 0.95 with an average of 0.87. Figure \ref{fig:Discuss_Depth}B-D shows three snapshots of in-finger vision, the depth map estimated from the inpainted view, and the SSI map. Low SSI values are most prominent at the objects' edges, where depth estimation is inherently more challenging.

    \paragraph{Scene Segmentation} Furthermore, we explored the feasibility of multi-object segmentation on the inpainted scene using the latest Segment Anything Model 2 (SAM2) \citep{Ravi2024SAM2}. We tested on the same scene described in the depth estimation task. 

    \begin{figure}[!h]
        \centering
        \includegraphics[width=1\linewidth]{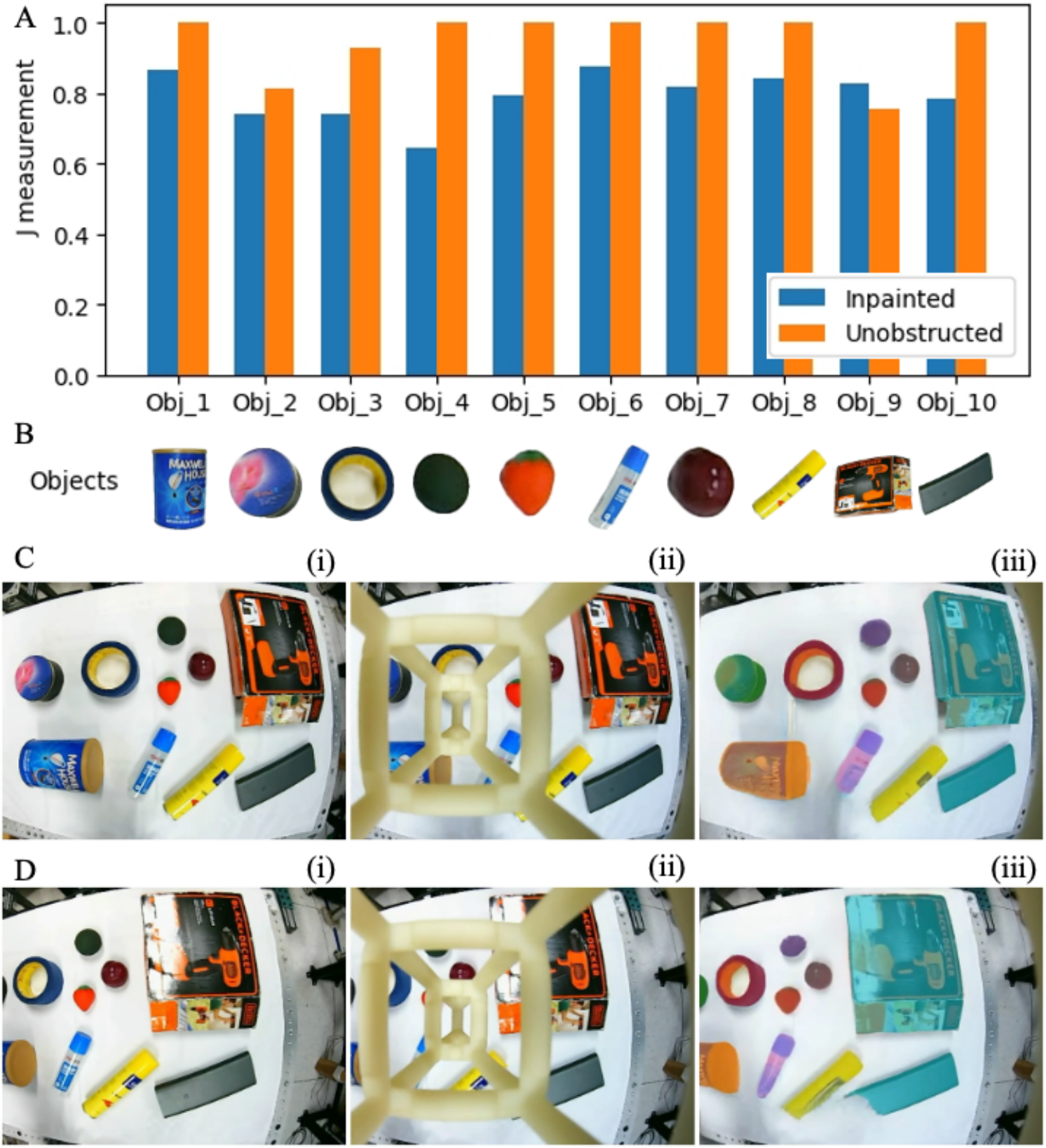}
        \caption{
            \textbf{Segmentation by seeing through the finger.}
            (A) The performance comparison of SAM2 on the inpainted and the unobstructed view of ten objects, measured by intersection over Union (Jaccard index);
            (B) Ten objects with various shapes and colors: milk powder can, tape, rubber ball, strawberry, glue, plum, glue stick, power drill with a packaging box, and drill toolkit;
            (C) Snapshot of the unobstructed and in-finger vision and the resultant segmentation of the inpainted scene;
            (D) A different snapshot similar to (C).
        }
        \label{fig:Discuss_Segmentation}
    \end{figure}
    
    As shown in Fig. \ref{fig:Discuss_Segmentation}A, segmentation on the inpainted view achieves satisfactory performance with J measurement ranging between 0.64 and 0.87, which is 16\% lower than the performance of segmentation on the unobstructed view on average. An interesting observation emerged during the segmentation of object 9 (a power drill with a packaging box): the inpainted view performed better than the unobstructed view. Further frame-by-frame examination shows that the worst segmentation in the unobstructed view was caused by the intense surface reflection of the drill's packaging box, and in some frames, SAM2 failed to segment object 9. On the contrary, SAM2 successfully tracked object nine in the inpainted view. The snapshots of the unobstructed and in-finger vision and segmentations of the inpainted scene at two different times are shown in Figs.~\ref{fig:Discuss_Segmentation}C\&D.

\subsection*{Multimodal \& Markerless VBSeeThruP}

    An intuitive understanding of the third research problem is whether we can utilize in-finger vision with a soft finger to achieve high-performing, multimodal perception in a markerless manner. Similar capabilities have been widely studied in the past literature, but the use of artificial markers remains a challenge among current solutions \citep{Yamaguchi2016Combining, Lambeta2020DIGIT, Lepora2021TacTip, Fan2024ViTacTip}. In this study, a by-product while addressing the first research problem is the real-time, pixel-wise segmentation of the See-Thru-Network's mask \(\tensor*[]{\mathbf{M}}{^{\text{Seg}}_{t}}\) throughout the implementation of the VBSeeThruP architecture. Visual features from this mask provide robust sensory input for inferring FT information, once a calibration dataset is obtained against commercial FT sensors for model training. 
    
    While simpler neural networks are reasonable, we adopted a generative approach in this study and used the Supervised Variational Autoencoder (SVAE) to address this problem. The latent variables obtained from SVAE are a powerful tool for inferring various sensory modalities, requiring minimal disruption to the overall computation. Besides the 6D FT, we demonstrate a simple method for inferring the contact event on the soft finger. Recent research also shows promising results in proprioceptive shape reconstruction \citep{Guo2024ProprioceptiveState}, contact point estimation \citep{Guo2024ReconstructingSoft}, and tactile perception using in-finger vision \citep{Liu2024ProprioceptiveLearning} with the soft polyhedral network used in this study, which could be further implemented using the deformation mask tracked in this study.

    \paragraph*{Contact Event Detection} In previous work where the complex deformations of SPN are physically encoded into a much lower 6D space of AruCo Marker pose, the contact location information has been lost \citep{Liu2024ProprioceptiveLearning}. With a much higher dimension of 32 in the SVAE framework, we hypothesize that such information could be maintained in the latent space to a certain degree. A preliminary test was conducted by categorizing the training samples into three contact events: no contact, contact at the finger tip, and contact in the middle region. We trained a contact event classifier using the SVAE framework shown in Fig. \ref{fig:Result_SVAE}, which can be formulated by
    \begin{equation}\label{eq:ContactEvent}
        \begin{aligned}
            \{\tensor*[]{\mathbf{Event}}{_{\text{Contact}}},\tensor*[]{\mathbf{M}}{^{\text{Gen}}}\} &= f_3(\tensor*[^{\text{STN}}_{\text{Dynamic}}]{\mathbf{M}}{^{\text{Seg}}}, \tensor*[]{\mathbf{M}}{^{\text{Seg}}_{t=0}}),\\
            f_3 &= \text{SVAE}.
        \end{aligned}
    \end{equation}

    \begin{figure}
        \centering
        \includegraphics[width=0.6\linewidth]{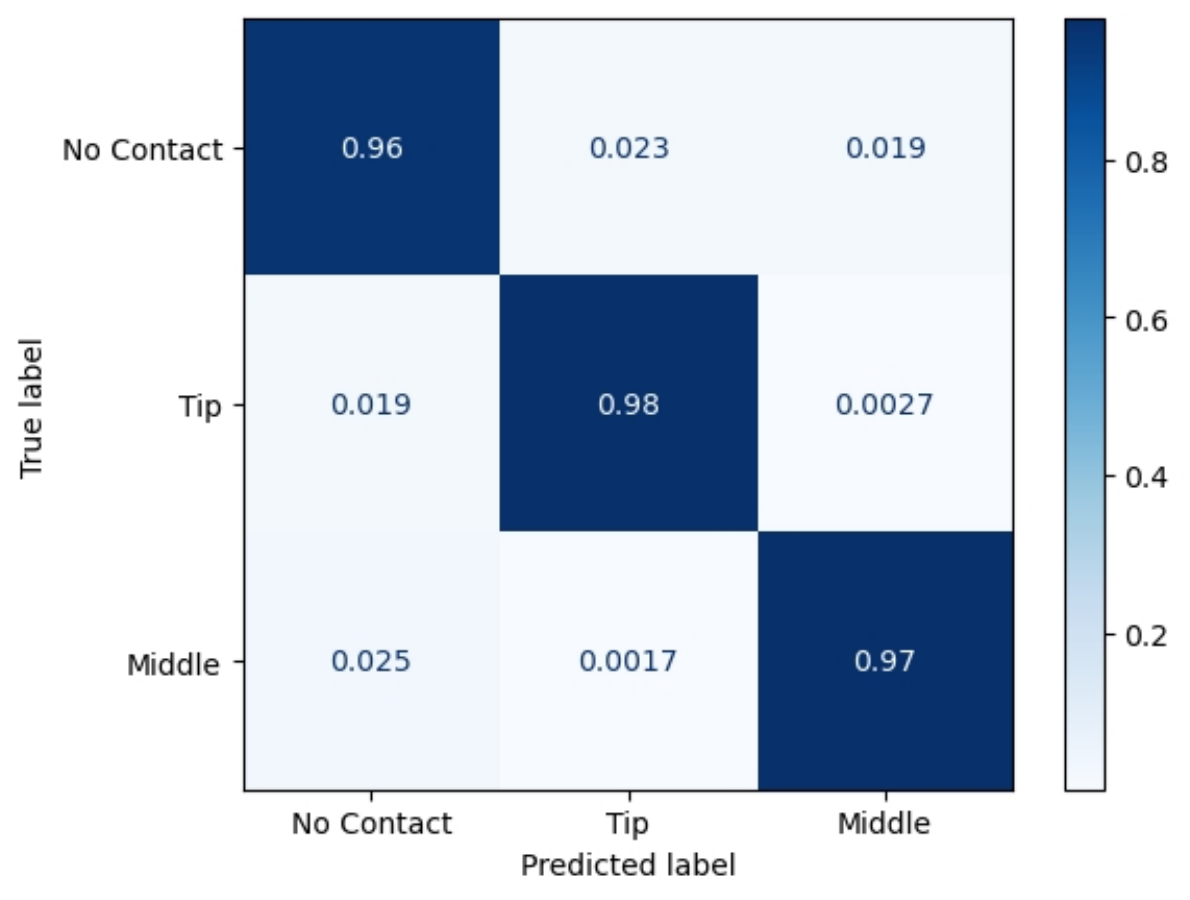}
        \caption{
            \textbf{Confusion matrix of contact event detection.} The tip and middle labels refer to the top and middle regions of the soft polyhedral network during contact.
        }
        \label{fig:Discuss_ContactEvent}
    \end{figure}

    Figure \ref{fig:Discuss_ContactEvent} shows the confusion matrix of the classifier evaluated on the validation dataset of 8,000 samples. The accuracies for all three contact events are above 97\%. Given that the classifier only takes the SPN's mask as input, which filters out all see-through visual information, the results confirm our hypothesis that the soft touch represented by the latent vectors contains information such as force/torque and contact location. A real-time demonstration is included in the supplementary video. The binary classification of contact locations is partly constrained by the data-collection process, which did not record precise locations. The granularity could be further improved by using denser labels or by converting it into a regression task.

\section{Conclusion, Limitations, and Future Directions}
\label{sec:Conclude}

    Inspired by sensory substitution from biological agents, this study presents a Vision-based See-Through Perception (VBSeeThruP) approach that reasons across different image layers to extract visual- and contact-based information from a single camera input. By placing a See-Thru-Network in front of a single in-finger camera, two layers of information can be obtained: one at a closer range for the soft interface, awaiting contact interactions, and another at a distance but occluded, about the scene. Contact-based perception is substituted markerlessly by reasoning about the masked deformation, facilitating reactive see-through grasping. 

    Despite encouraging results, there are notable constraints. First, while XMem runs at 30 Hz, it remains significantly slower than the camera's potential 330 Hz frame rate, and the overall pipeline still needs optimization for high-speed robotic tasks. Second, mislabeled detections sometimes occur because we did not fine-tune object detection for specific targets. Third, the learned 6D force/torque model exhibits a systematic shift when applied to new fingers with slightly different initial templates. Although subtracting this offset mitigates the error, more effective data augmentation strategies could further improve generalization. Finally, larger scenes often appear blurry through the STN, limiting the range of reliable visual perception.

    Several directions remain open. Promising avenues include underwater perception and manipulation \citep{Oussama2016OceanOne}, embodied robot learning with multimodal feedback that integrates SPN's omni-directional adaptability and precise 6D force/torque sensing \citep{Chi2024UniversalManipulation}, and extending the single-camera approach to stereo vision for improved 3D reconstruction or 6D pose estimation.

\section*{Acknowledgement}

    This work was partly supported by the National Natural Science Foundation of China under Grant 62473189.

\section*{Data Availability}

    All data and codes related to this work are hosted on the following project website: \url{https://github.com/ancorasir/SeeThruFinger}.



\bibliographystyle{elsarticle-harv} 
\bibliography{References}
\end{document}